\documentclass[runningheads]{llncs}
\usepackage{graphicx}

\usepackage{pifont}%
\usepackage{multirow}
\usepackage{caption}
\usepackage{subcaption}
\usepackage{graphicx}
\usepackage{amssymb}
\usepackage{sidecap}
\usepackage{wrapfig,lipsum,booktabs}
\usepackage{comment}
\usepackage{ulem}
\usepackage{amsmath}
\usepackage{float}
\usepackage{placeins}

\usepackage[pagebackref,breaklinks,colorlinks]{hyperref}

\usepackage[capitalize]{cleveref}
\crefname{section}{Sec.}{Secs.}
\Crefname{section}{Section}{Sections}
\Crefname{table}{Table}{Tables}
\crefname{table}{Tab.}{Tabs.}

\usepackage[accsupp]{axessibility}  

\usepackage[width=122mm,left=12mm,paperwidth=146mm,height=193mm,top=12mm,paperheight=217mm]{geometry}

\usepackage{amsmath,amsfonts,bm}

\newcommand{\model}{\text{BUTD-DETR }}
\newcommand{\Det}{\texttt{det}}
\newcommand{\GT}{\texttt{GT}}

\def\eqref#1{equation~\ref{#1}}

\def\1{\bm{1}}

\newcommand{\test}{\mathcal{D_{\mathrm{test}}}}

\DeclareMathAlphabet{\mathsfit}{\encodingdefault}{\sfdefault}{m}{sl}
\SetMathAlphabet{\mathsfit}{bold}{\encodingdefault}{\sfdefault}{bx}{n}

\usepackage{hyperref}
\usepackage{url}

\newcommand\blfootnote[1]{%
  \begingroup
  \renewcommand\thefootnote{}\footnote{#1}%
  \addtocounter{footnote}{-1}%
  \endgroup
}

\begin{document}



\pagestyle{headings}
\mainmatter
\def\ECCVSubNumber{4656} 

\title{Bottom Up Top Down Detection Transformers for Language Grounding in Images and Point Clouds}

\titlerunning{Bottom Up Top Down Detection Transformers} 
\authorrunning{Jain and Gkanatsios et al.}
\author{
		Ayush Jain$^{\dagger1}$, Nikolaos Gkanatsios$^{\dagger1}$, Ishita Mediratta$^{\mathsection2}$, and Katerina Fragkiadaki$^1$
	}
\institute{$^1$Carnegie Mellon University, $^2$Meta AI}
\newcommand{\fix}{\marginpar{FIX}}
\newcommand{\new}{\marginpar{NEW}}
\maketitle
\blfootnote{$^{\dagger}$Equal contribution, order decided by \texttt{np.random.rand}}
\blfootnote{$^{\mathsection}$Work done during an internship at CMU}

\begin{abstract}
Most models tasked to ground referential utterances in 2D and 3D scenes learn to select the referred object from a pool of object proposals provided by a pre-trained detector. 
This  is limiting because an utterance may refer to visual entities at various levels of granularity, such as the chair, the leg of the chair, or the tip of the front leg of the chair, which may be missed by the  detector. 
 We propose  a language  grounding model that attends on the referential utterance and on the  object proposal pool computed  from a pre-trained detector to  decode referenced objects with a detection head, without  selecting them from the pool. In this way, it  is helped by powerful pre-trained object detectors without being restricted by their misses.   
We call our model Bottom Up Top Down DEtection TRansformers (BUTD-DETR) because it uses both language guidance (top down) and  objectness guidance (bottom-up) to ground referential utterances in images and point clouds.   
 Moreover, \model{} casts object detection as referential grounding  and uses object labels  as language prompts to be grounded in the visual scene,  augmenting supervision for the referential grounding task in this way. 
 The proposed model sets a new state-of-the-art across 
popular 3D language grounding benchmarks with significant performance gains over previous 3D approaches (12.6\% on SR3D, 11.6\% on NR3D and 6.3\% on ScanRefer). 
When applied in 2D images, it performs  on par with the previous state of the art. 
We ablate the design choices  of our model and  quantify their contribution to performance.
Our code and checkpoints  can be found at the project website  \url{https://butd-detr.github.io}. 
\end{abstract}

\section{Introduction}

\begin{figure*}[h!]
	\centering
	\includegraphics[width=\textwidth]{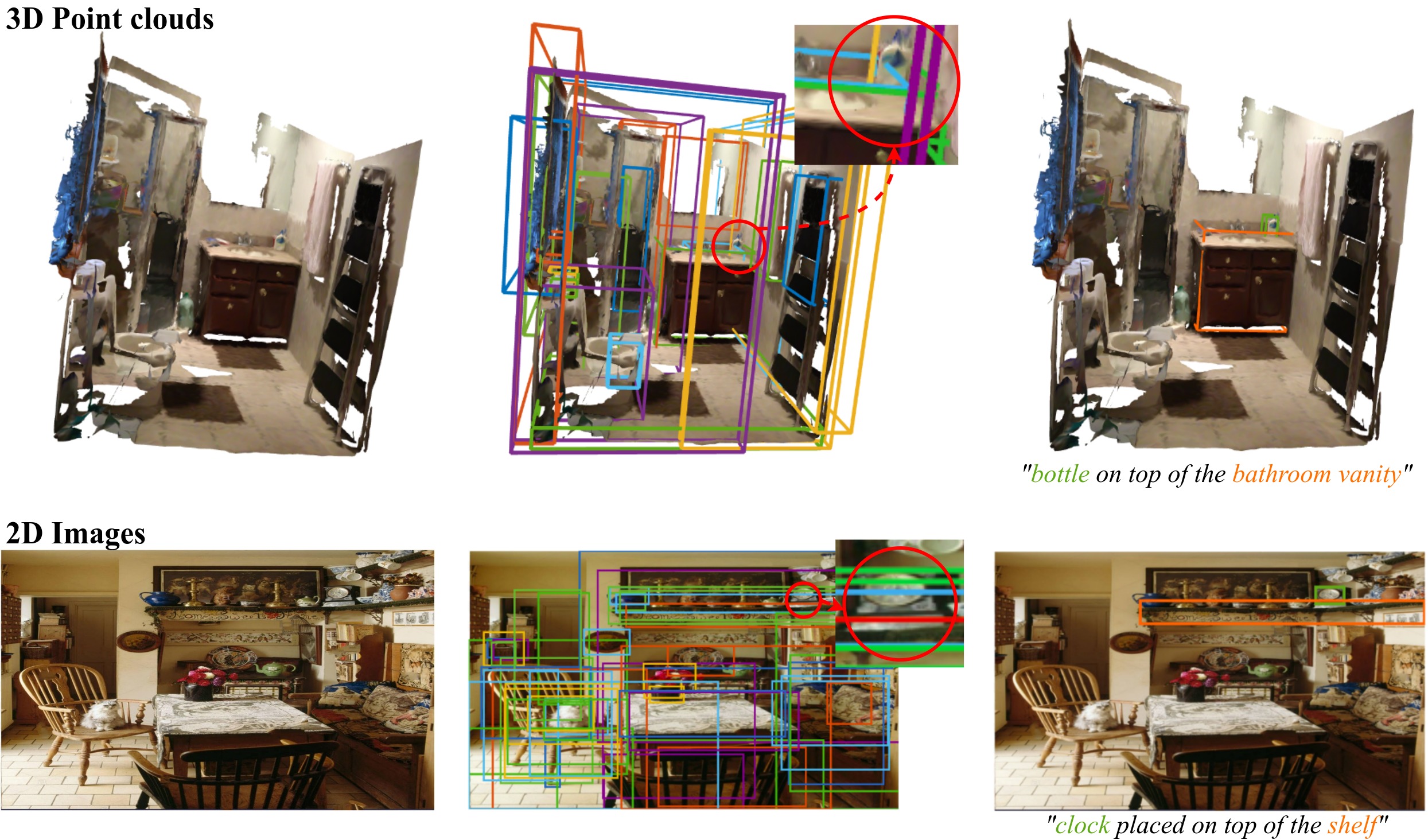}
	\caption{\textbf{Language-modulated 3D (\textit{top}) and 2D (\textit{bottom}) detection
	with BUTD-DETR.} \textit{Middle:} 
	State-of-the-art object detectors 
	often fail to localize small, occluded or rare objects (here they miss the clock on the shelf and the bottle on the cabinet). 
	\textit{Right:}
	Language-driven and objectness-driven attention in \model modulates the visual processing  depending on the referential expression while taking into account salient, bottom-up detected objects, and correctly localizes all referenced objects. 
	}
	\label{fig:overview}
\end{figure*}

Language-directed attention  helps us localize objects that  our ``bottom-up",  task-agnostic perception may miss. 
Consider Fig. \ref{fig:overview}. 
The  utterance  \textit{``bottle on top of the bathroom vanity"} suffices to  direct our attention to the reference object, even though it is far from salient.  
Language-directed perception  adapts the visual processing of the input scene according to the   utterance. 
Object detectors instead apply the same computation in each  scene, which can miss task-relevant objects.

Most existing language grounding models use object proposal bottlenecks: they select the referenced object from a pool of object proposals provided by the pre-trained object detector  \cite{fang2015captions,densecap,karpathy2015deep2,Fukui2016MultimodalCB,modularreferential}. This means they cannot recover objects or parts that a bottom-up detector misses. This is limiting since small, occluded, or rare objects are hard to detect without task-driven guidance. For example, in Figure~\ref{fig:overview} middle,  state-of-the-art 2D \cite{ren2015faster} and 3D \cite{Liu2021GroupFree3O} detectors  miss the clock on the shelf and the bottle on the bathroom vanity, respectively.

Recently, Kamath et al. \cite{Kamath2021MDETRM} introduced  MDETR, a  language grounding model for 2D images that decodes object boxes using a DETR \cite{Carion2020EndtoEndOD} detection head and  aligns them to the relevant  spans in the input utterance, it does not select the answer from a box proposal pool. The visual computation is modulated based on the input utterance through   several layers of self-attention on a concatenation of language and visual features. 
MDETR  achieves big leaps in performance in 2D language grounding over previous box-bottlenecked methods. 


 
 We propose a model for grounding referential utterances in 3D and 2D visual scenes that builds upon MDETR,  which we call \underline{BUTD}-DETR (pronounced Beauty-DETR),  as it uses both box  proposals, obtained by a pre-trained detector  ``\underline{b}ottom-\underline{u}p"  and ``\underline{t}op-\underline{d}own" guidance from the  language utterance, to localize the relevant objects in the scene. 
 \model{}  
 uses box proposals obtained by a pre-trained detector  as an  additional input stream to attend on; however, it is not  box-bottlenecked  and   still decodes objects with a detection head, instead of selecting them from the input box stream. Current object detectors  provide a noisy tokenization of the  input visual scene that, as our experiments show, is a useful cue to attend on for multimodal reasoning.  
 Second, \model{} augments  grounding annotations   by configuring  annotations for object detection as detection prompts to be grounded in visual scenes.  A detection prompt is a list of object category labels, e.g., \textit{``Chair. Door. Person. Bed."}.
 We train the model to ground detection prompts by localizing the labels that are present in the image and learn to discard labels that are mentioned but do not correspond to any objects in the scene. 
 Third, \model{} considers improved bounding box - word span alignment losses that reduce  noise during alignment of object boxes to noun phrases in the referential utterance.  

We test $\model$ on the 3D benchmarks of \cite{Achlioptas2020ReferIt3DNL,Chen2020ScanRefer3O} and 2D benchmarks of \cite{Kazemzadeh2014ReferItGameRT,Yu2016ModelingCI}. In 3D point clouds,  we set new state-of-the-art in the two benchmarks of Referit3D \cite{Achlioptas2020ReferIt3DNL} and ScanRefer \cite{Chen2020ScanRefer3O} and report significant performance boosts over all prior methods (12.6\% on SR3D, 11.6\% on NR3D and 6.3\% on ScanRefer), as well as over a direct MDETR-3D implementation of ours that does not use a box proposal stream or detection prompts during training.
In 2D images, our model  obtains competitive performance with MDETR on RefCOCO, RefCOCO+ and Flickr30k, and requires less than half of the GPU training time due to the cheaper deformable attention in the visual stream. We ablate each of the design choices of the model to quantify their contribution to performance. 

In summary, our contributions are: 
\textbf{(i)} A model with SOTA performance across both 2D and 3D scenes with minor changes showing that modulated detection in 2D images can also work in 3D point clouds with appropriate visual encoder and decoder modifications.   
\textbf{(ii)} Augmenting supervision with detection prompts, attention on an additional input box stream and improved bounding box - word span alignment losses.
 \textbf{(iii)} Extensive  ablations to quantify the contribution of different components of our model. 
We make our code publicly available at
\url{https://butd-detr.github.io}.

\section{Related work}

\subsubsection{Object detection with transformers}
Object detectors are trained  
to localize all instances of a closed set of object category labels in images and 3D point-clouds. While earlier architectures  pool features within proposed boxes to decode objects and classify them into categories 
\cite{DBLP:journals/corr/HeGDG17,DBLP:journals/corr/LiuAESR15,Redmon2016YouOL},  recent methods pioneered by DETR \cite{Carion2020EndtoEndOD} use transformer architectures where a set of object query vectors attend to the scene and among themselves to decode object boxes and their labels. DETR suffers from the quadratic cost of within  image features  self attention.  D(eformable)-DETR \cite{Zhu2021DeformableDD} proposes  deformable attention, a locally adaptive kernel that is predicted directly in each pixel location without attention to other pixel locations, thus saving the quadratic cost of pixel-to-pixel attention.
Our model builds upon deformable attention  for feature extraction from RGB images. 
\cite{Liu2021GroupFree3O,Misra2021AnET} extend detection transformers to 3D point cloud input.

\subsubsection{2D referential language grounding}
Referential language grounding \cite{Kazemzadeh2014ReferItGameRT} is the task  of  localizing the object(s) referenced in a language  utterance. 
Most 2D language grounding models  obtain  sets of object proposals using  pre-trained  object detectors and  the original image is discarded upon extraction of the object proposals
\cite{fang2015captions,densecap,karpathy2015deep2,Fukui2016MultimodalCB,modularreferential}. Many of these approaches use multiple layers of attention to fuse information across both, the extracted boxes and language utterance \cite{Lu2019ViLBERTPT,Chen2020UNITERUI,Yang2021SAT2S}. Recently, a few approaches directly regress the target bounding box without using pre-trained object proposals. In \cite{chen2018real} language and visual features cross-attend
and are concatenated to predict the box of the referential object. Yang et al. \cite{yang2019fast} extends the YOLO detector \cite{Redmon2016YouOL} to referential grounding by channel-wise concatenating language, visual and spatial feature maps and then regressing a single box using the YOLO box prediction head.
\cite{sadhu2019zero} performs a fusion similar to \cite{yang2019fast}, then selects a single box from a set of anchor boxes and predicts a deformation of it, much like the Faster-RCNN object detector \cite{ren2015faster}. 
While previous approaches encode the whole text input into a single feature vector, \cite{yang2020improving} further improves performance by recursively attending on different parts of the referential utterance.
Lastly, \cite{deng2021transvg} encodes the image and utterance with within- and cross-modality transformers, and a special learnable token regresses a single box. In contrast to our method, all these works predict a single bounding box per image-utterance pair. 
Our work builds upon MDETR of Kamath et al. \cite{Kamath2021MDETRM} that modulates visual processing through attention to the input language utterance and decodes objects from queries similar to DETR, without selecting from a pool of proposals. Both our method and MDETR can predict multiple instances being referred to, as well as ground intermediate noun phrases. 
Concurrent to our work, GLIP \cite{Li2021GroundedLP} shows that adding supervision from detection annotations can improve 2D referential grounding. Our work independently confirms this hypothesis in 2D and also shows its applicability on the 3D domain.

    \subsubsection{3D referential language grounding} has only recently gained popularity \cite{Chen2020ScanRefer3O,Achlioptas2020ReferIt3DNL}. To the best of our knowledge, all related approaches are box-bottlenecked: they extract 3D object proposals and select one as their answer. Their pipeline can be decomposed into three main steps: i) Representation of object boxes as point features \cite{Yang2021SAT2S}, segmentation masks \cite{Yuan2021InstanceReferCH} or pure spatial/categorical features \cite{Roh2021LanguageReferSM}. ii) Encoding of language utterance using word embeddings \cite{Yang2021SAT2S,Roh2021LanguageReferSM} and/or scene graphs \cite{Feng2021FreeformDG}. iii) Fusion of the two modalities and scoring of each proposal using graph networks \cite{Huang2021TextGuidedGN} or Transformers \cite{Yang2021SAT2S}. Most of these works also employ domain-specific design choices by explicitly encoding pairwise relationships  \cite{Huang2021TextGuidedGN,he2021TransRefer3DEA,Yuan2021InstanceReferCH} or by relying on heuristics, such as restricting attention to be local \cite{Zhao_2021_ICCV,Yuan2021InstanceReferCH} and ignoring input modalities \cite{Roh2021LanguageReferSM}. Such design prevents those architectures from being applicable to both the 3D and 2D domains simultaneously.
    
    Due to the inferior performance of 3D object detectors in comparison to their 2D counterparts, popular benchmarks for 3D language grounding, such as Referit3D \cite{Achlioptas2020ReferIt3DNL} provide access to ground-truth object boxes at test time. The proposed $\model$ is the first 3D language grounding model that is evaluated on this benchmark without access to  oracle 3D object boxes. 
 
\section{Method}

We first describe MDETR \cite{Kamath2021MDETRM} in Section \ref{sec:background}. Then, we present BUTD-DETR's architecture in Section \ref{sec:architecture}, supervision augmentation  with detection prompts in Section \ref{sec:prompts} and its training objectives in Section \ref{sec:objectives}.

\subsection{Background: MDETR} \label{sec:background}
MDETR is a  2D language grounding model that takes  a referential utterance and an RGB image as input and  localises in the image all objects mentioned in the utterance. MDETR encodes the image  with a convolutional network \cite{he2016deep} and the language utterance with a RoBERTa encoder \cite{Liu2019RoBERTaAR}. It then fuses information across   the language and visual features through multiple layers of self-attention on the concatenated visual and language feature sequences.   
In MDETR's decoder, a set of query vectors  iteratively attend to the contextualized visual features and self-attend to one another, similar to the DETR's \cite{Carion2020EndtoEndOD} decoder.  Finally, each query  decodes a bounding box and a confidence score over each word in the input utterance, which associates the box to a text span.

The predicted boxes are assigned to ground-truth ones using a  Hungarian matching, similar to  \cite{Carion2020EndtoEndOD}. 
Upon matching, 
the following losses are  computed:
\begin{itemize}
    \item A bounding box  loss between predicted boxes and  the corresponding ground-truth ones. This is a combination of L1 and generalized IoU \cite{Rezatofighi2019GeneralizedIO} losses.
    
    \item A soft token prediction loss. A query matched to a ground-truth box is trained to decode  a uniform distribution over the language token positions that refer to that object. Queries not matched to ground-truth targets are trained to predict a no-object label. 
    
    \item Two contrastive losses between query and language token features. The first one, called \textit{object contrastive loss}, pulls an object  query's features closer to the  features of the corresponding ground-truth span's word tokens, and further than   all other tokens. The second one, called \textit{token contrastive loss}, pulls the features of a ground-truth span's token closer to the corresponding object query features, and further than all other queries.
\end{itemize}

\begin{figure*}[t]
\begin{center}
    \includegraphics[width=\textwidth]{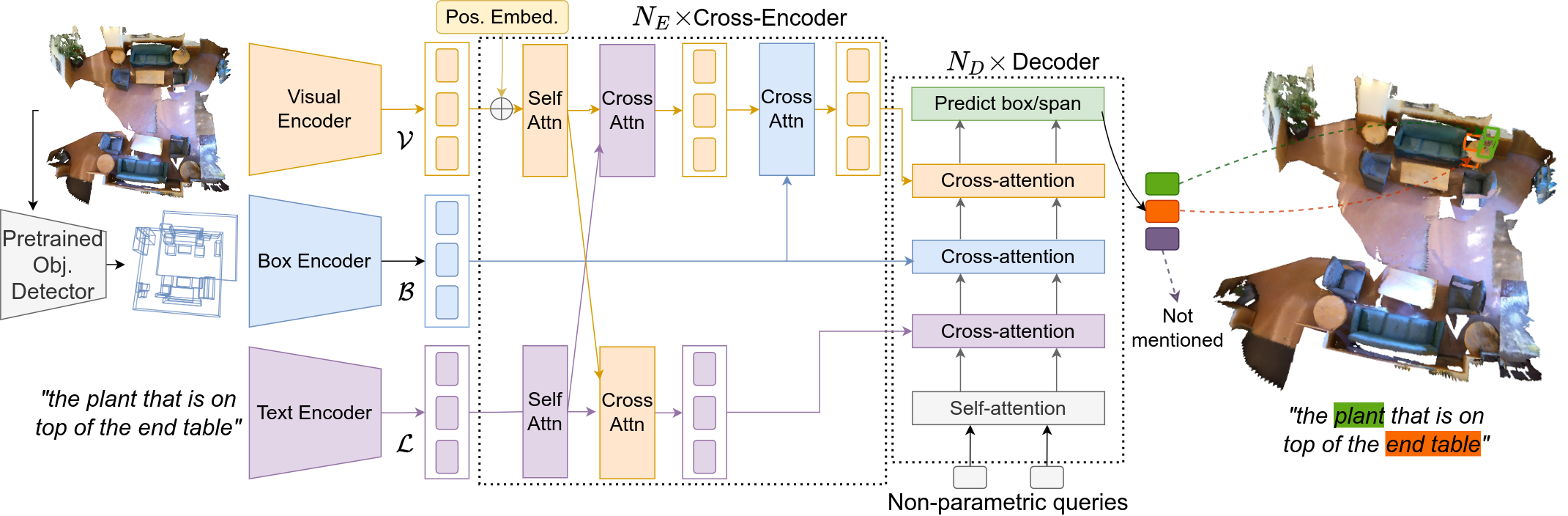}
\end{center}
\caption{\textbf{BUTD-DETR architecture.} 
Given a  visual scene  and a referential utterance,  the model localizes  all  object instances  mentioned in the utterance.  
A pre-trained object detector extracts object box proposals. 
The visual scene features, the language utterance and the labelled box proposals 
are encoded into corresponding sequences of visual, word and box tokens  using  visual, language and box 
encoders, respectively. 
The three streams cross-attend and finally decode boxes and corresponding   spans in the language utterance that each decoded box refers to. We visualize here the model operating on a 3D point cloud; an analogous architecture is used for 2D image grounding. 
}
\label{fig:arch_det}
\end{figure*}

\subsection{Bottom-up Top-down DETR (BUTD-DETR)} \label{sec:architecture} 
The  architecture of $\model$  is illustrated in Figure ~\ref{fig:arch_det}. 
Given a referential language utterance, e.g., ``find the plant that is on top of the end table'' and a visual scene, which can be a 3D point cloud or a 2D image,  $\model$ is trained to localize all objects mentioned in the utterance. In the previous example, we expect one box for the ``plant'' and one for the ``end table''. 
The model  attends across image/point cloud, language and box proposal streams, then decodes the relevant objects  and aligns them to input language spans.

\subsubsection{Within-modality encoder}
In 2D, we encode an  RGB image  using a pre-trained ResNet101  backbone \cite{He2016DeepRL}. The 2D appearance visual features are added to 2D Fourier positional encodings, same as in \cite{Zhu2021DeformableDD,Jaegle2021PerceiverGP}. 
In 3D,  we encode a 3D point cloud using a PointNet++ backbone \cite{Qi2017PointNetDH}. The 3D point visual features are added to  learnable 3D positional encodings, same as in  \cite{Liu2021GroupFree3O}: we pass the coordinates of the points through a small multilayer perceptron (MLP). 
Let $\mathcal{V} \in \mathbb{R}^{n_v \times c_v}$ denote  the visual token sequence, 
where $n_v$ is the number of visual tokens and $c_v$ is the number of visual feature channels.

The words of the input utterance are encoded using a pre-trained RoBERTa \cite{Liu2019RoBERTaAR} backbone. 
Let $\mathcal{L} \in \mathbb{R}^{n_\ell \times c_\ell}$ denote  the word token sequence.

A pre-trained detector is used to obtain 2D or 3D object box proposals.  Following prior literature, we use Faster-RCNN \cite{ren2015faster} for RGB images, pre-trained on 1601 object categories of Visual Genome \cite{Krishna2016VisualGC}, and Group-Free detector \cite{Liu2021GroupFree3O} for 3D point clouds pre-trained on a vocabulary of 485 object categories on ScanNet \cite{Dai2017ScanNetR3}. The detected box proposals that surpass a confidence threshold 
are encoded using a box proposal encoder, by mapping their spatial coordinates and categorical class information to an embedding vector each, and concatenating them to form an object proposal token. We use a pre-trained and frozen RoBERTa \cite{Liu2019RoBERTaAR} backbone  to encode the semantic categories of proposed boxes. 
Let  $\mathcal{O} \in \mathbb{R}^{n_o \times c_o}$ denote the object token sequence. 

The 3D detector is trained on ScanNet and all 3D benchmarks we use are also ScanNet-based. This creates a discrepancy in the quality of the detector's predictions between train and test time, as it is far more accurate on the training set. As a result, we find that \model tends to rely on the detector at training time and generalizes less at test time, where the detector's predictions are much noisier. To mitigate this, we randomly replace 30\% of the detected boxes at training time with random ones. This augmentation leads to stronger generalization when the detector fails to locate the target object. Note that this is not the case in 2D, where the detector is trained on a different dataset.

All visual, word and box proposal tokens are mapped using (different per modality) MLPs to same-length feature vectors.


\subsubsection{Cross-modality Encoder} The visual,  language  and box proposals, interact through a 
sequence of 
$N_E$ 
cross-attention layers.   
In each encoding layer, visual  and language tokens cross-attend to one another and are updated using standard key-value attention.  
Then, the resulting language-conditioned visual tokens attend to the box proposal tokens. 
We use standard attention for both streams in 3D and deformable attention \cite{Zhu2021DeformableDD} for the visual stream in 2D. 

In contrast to MDETR, \model keeps visual, language and box stream separate in the encoder instead of concatenating them. This enables us to employ deformable attention \cite{Zhu2021DeformableDD} in self and cross attention layers involving the visual stream in 2D domain. Deformable attention involves computing bilinearly interpolated features which is expensive and non-robust in discontinous and sparse modalities like pointclouds, hence we use vanilla attention in 3D (for more details see supplementary).
In our experiments, we show that concatenation versus keeping separate streams performs similarly in 3D referential grounding. 





\subsubsection{Decoder}

\model{} decodes objects from contextualized features using non-parametric queries in both 2D and 3D, similar to \cite{Zhu2021DeformableDD,Liu2021GroupFree3O}. Non-parametric queries are predicted  by visual tokens from the current scene, in contrast to parametric queries used in DETR  \cite{Carion2020EndtoEndOD} and MDETR \cite{Kamath2021MDETRM} that correspond to a learned  set of vectors  shared across all scenes. Specifically, the contextualized visual tokens from the last multi-modality encoding layer  predict confidence scores, one per visual token. The top-$K$ highest scoring tokens are each fed into an MLP to predict a vector which stands for an \textit{object query}, i.e., a vector that will decode a box center and size relative to the location of the corresponding visual token, similar to D-DETR  \cite{Zhu2021DeformableDD}.   
The   query vectors are updated in a residual manner through $N_D$ decoder layers. In each decoder layer,  we employ four types of attention operations. First, the  queries self-attend to one another to contextually refine their estimates. Second, they attend to the contextualized word embeddings  to condition on the language utterance.  Next, they attend to the box proposal tokens and then in the image or point visual tokens. 
At the end of each decoding layer, there is a prediction head that predicts a box center displacement, height and width vector, and a token span for each object query that localizes the corresponding object box and aligns it with the language input. 
We refer the reader to our supplementary file for more implementation details.

\begin{figure*}[t]
\begin{center}
    \includegraphics[width=\textwidth]{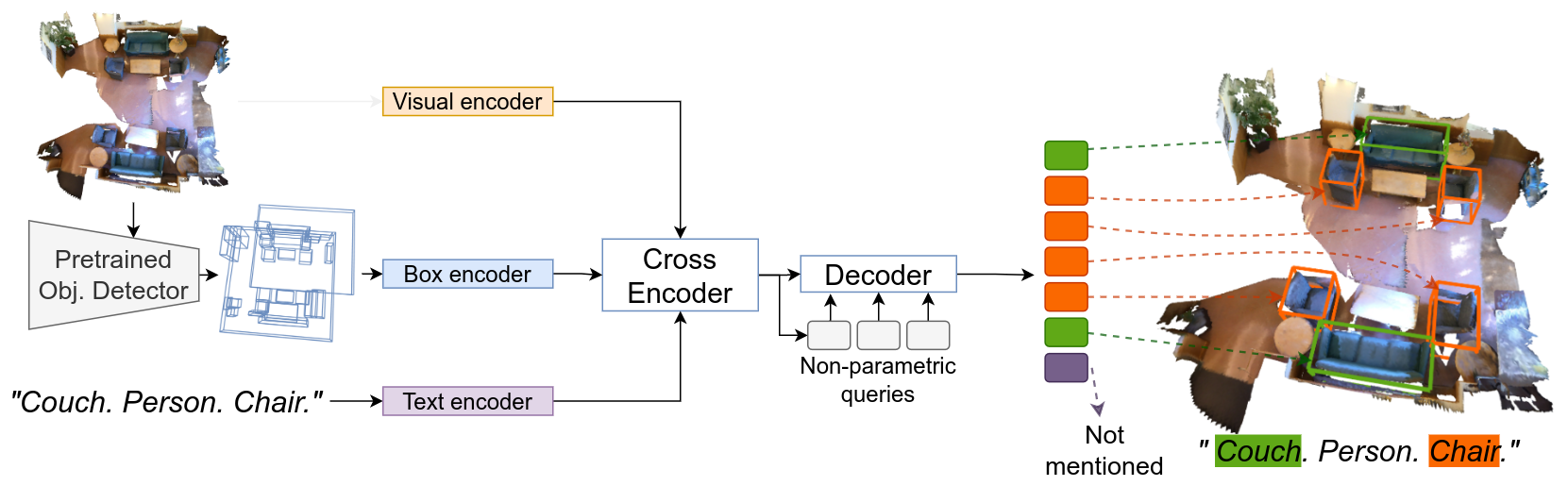}
\end{center}
\caption{\textbf{Augmenting referential grounding supervision  with detection prompts.} A detection prompt is constructed by sequencing sampled object category labels (here \textit{couch}, \textit{person} and \textit{chair}). The task is to localize all instances of mentioned objects and associate them with the correct span in the prompt. 50\% of the sampled labels are negative, i.e., they have no corresponding object instance in the scene. The model learns not to associate these spans with predicted boxes.}
\label{fig:prompt}
\end{figure*}

\subsection{Augmenting supervision  with detection prompts} \label{sec:prompts} 
Object detection is an instance of referential language grounding in which the utterance is a single word, namely, the object category label. 
Language grounding models have effectively combined  supervision across  referential grounding, caption description and question answering tasks \cite{Lu2019ViLBERTPT,Lu202012in1MV}, which is an important factor for  their success.   Object detection annotations have not been considered so far as candidates for such co-training. 



We cast object detection as  grounding of detection prompts, namely, referential utterances comprised of a list of object category labels, as shown in Figure \ref{fig:prompt}.  
Specifically, given the detector's vocabulary  of object category labels, we randomly sample a fixed number of them---some  appear in the visual scene and some do not---and  generate synthetic utterances by sequencing the sampled  labels, e.g.,  \textit{``Couch. Person. Chair. Fridge.''}, we call them detection prompts.  We treat these prompts as referential utterances to be grounded: the task is to localize \textit{all} object instances of the category labels mentioned in the prompt if they appear in the scene. The sampling of negative  category labels (labels for which there are no object instances present) operates as negative training:  the model is trained to not match any boxes to the negative category labels. Further details on this negative training can be found in the supplementary. 


\subsection{Supervision objectives} \label{sec:objectives}
We  supervise the outputs of all  prediction heads in each layer of the decoder. We follow MDETR \cite{Kamath2021MDETRM} in using Hungarian matching to assign a subset of object queries to the ground-truth object boxes and then compute the bounding box, soft token prediction and contrastive losses. 
Our bounding box and soft token prediction losses are identical to MDETR's. 
However, we notice that MDETR's contrastive losses do not compare all object queries and word tokens symmetrically. Specifically, the object contrastive loss supervises only the object queries that are matched to a ground-truth object box. On the other hand, the token contrastive loss includes only the tokens that belong to positive spans, namely, noun phrases with corresponding object instances in the scene. As a result, object queries not matched to any ground-truth object box are not pulled far from non-ground-truth text spans, which means at inference object queries can be close to negative spans.  
We find this asymmetry to hurt performance, as we show in our experiments.

To address this, we propose a symmetric alternative where the similarities between all object queries and language tokens are considered. 
We append the span ``not-mentioned'' to all input utterances. This acts as the ground-truth text span for all object queries that are not assigned to any of the ground-truth objects. The object contrastive loss now supervises all queries and considers the similarities with all tokens. We empirically find that gathering unmatched queries to ``not mentioned'' is beneficial. This is similar in principle to the soft token prediction loss, where unmatched queries have to predict ``no object''. In fact, we find that this symmetric contrastive loss is sufficient for our model's supervision, but we observe that co-optimizing for soft token prediction results in faster convergence. 


\section{Experiments} \label{sec:exps}

We test $\model$ on grounding referential utterances  in 3D point clouds and 2D  images. 
Our experiments aim to answer the following questions: 
\begin{enumerate}
    \item  How does $\model$ perform compared to the   state-of-the-art in 3D and 2D language grounding?
     \item  How does $\model$ perform compared to a straightforward extension of the 2D state-of-the-art MDETR \cite{Kamath2021MDETRM} model in 3D?
    \item How much, if at all, attending to a bottom-up box proposal stream helps performance?
    \item How much, if at all, co-training for grounding detection prompts helps performance?
    \item How much, if at all, the proposed contrastive loss variant helps performance?
\end{enumerate}

\subsection{Language grounding in 3D point clouds}
We test 
\model on SR3D, NR3D \cite{Achlioptas2020ReferIt3DNL} and ScanRefer \cite{Chen2020ScanRefer3O} benchmarks. All three benchmarks contain pairs of 3D point clouds of indoor scenes from ScanNet \cite{Dai2017ScanNetR3} and corresponding referential utterances, and the task is to localize the objects referenced in the utterance. 
The utterances in SR3D are short and synthetic, e.g., \textit{``choose the couch that is underneath the picture"},
while utterances in 
NR3D and ScanRefer  are longer and more natural, e.g. \textit{``from the set of chairs against the wall, the chair farthest from the red wall, in the group of chairs that is closer to the red wall".}
For fair comparison against previous methods, we  train $\model$  separately on each of SR3D, NR3D and ScanRefer. We augment supervision in each of the three datasets with ScanNet detection prompts. SR3D provides annotations for all objects mentioned in the utterance, so during training we supervise localization of all objects mentioned. In NR3D and ScanRefer,  we use supervision for grounding only the referenced object. 



All existing models that have been tested in SR3D or NR3D benchmarks are box-bottlenecked, namely, they are trained to select the answer from a pool of box proposals.  They all use \textbf{ground-truth 3D object boxes (without category labels)} as the set of boxes to select from. 
We thus consider two evaluation setups: 
\begin{enumerate}
    \item $\Det$: where we re-train  previous models using their publicly available code and provide the same 3D box proposals we use in BUTD-DETR, obtained by the Group-Free 3D object detector \cite{Liu2021GroupFree3O} trained to detect 485 object categories in ScanNet (Section $\Det$ in Table \ref{table:3d_results}).
    \item $\GT$, where we use ground-truth 3D object boxes for our model and baseline (Section $\GT$ in Table~\ref{table:3d_results}). 
\end{enumerate}

Alongside previous  models, we also  compare our model against our implementation  of the MDETR model in 3D. This is similar to our model but without attention on a box stream, without co-training with detection prompts and with the original contrastive losses proposed by MDETR. We also replace MDETR's parametric object queries with non-parametric one ---similar to our model---since they have been shown to be crucial for good performance in 3D \cite{Liu2021GroupFree3O,Misra2021AnET}. We call this model MDETR-3D. For the sake of completeness, we do have a 3D version of MDETR that uses parametric queries in Table~\ref{table:ablation} and, as expected, it is significantly worse. MDETR does not use a pool of box proposals in any way and hence we cannot report results of MDETR-3D under \GT.

\begin{table*}[t]
    \centering
    \caption{\textbf{Results on language grounding in 3D point clouds.} We evaluate top-1 accuracy using ground-truth ($\GT$) or detected ($\Det$) boxes. $^{*}$ denotes method uses extra 2D image features. $^{\dagger}$ denotes evaluation with detected boxes using the authors' code and checkpoints. $^{\ddagger}$ denotes re-training using the authors' code. For \cite{Zhao_2021_ICCV}, we compare against their 3D-only version.}
    \resizebox{.98\textwidth}{!}
    {
    \begin{tabular}{|l||*{6}{c|}}
        \hline
        & \multicolumn{2}{c|}{SR3D} & NR3D & \multicolumn{2}{c|}{ScanRefer (Val. Set)}\\
        Method & Acc@0.25($\Det$) & Acc.($\GT$) & Acc@0.25($\Det$) & Acc@0.25($\Det$) & Acc@0.5($\Det$) \\
        \hline
        ReferIt3DNet \cite{Achlioptas2020ReferIt3DNL} & 27.7$^{\dagger}$ & 39.8 & 24.0$^{\dagger}$ & 26.4 & 16.9\\
        ScanRefer \cite{Chen2020ScanRefer3O} & - & - & - & 35.5 & 22.4\\
        TGNN \cite{Huang2021TextGuidedGN} & - & 45.0 & - & 37.4 & 29.7\\
        3DRefTransformer \cite{Abdelreheem20223DRefTransformerFO} & - & 47.0 & - & - & -\\
        InstanceRefer \cite{Yuan2021InstanceReferCH} & 31.5$^{\ddagger}$ & 48.0 & 29.9$^{\ddagger}$ & 40.2 & 32.9\\
        FFL-3DOG \cite{Feng2021FreeformDG} & - & - & - & 41.3 & 34.0 \\
        LanguageRefer \cite{Roh2021LanguageReferSM} & \underline{39.5}$^{\dagger}$ & 56.0 & 28.6$^{\dagger}$ & - & - \\
        3DVG-Transformer \cite{Zhao_2021_ICCV} & - & 51.4 & - & \underline{45.9} & \underline{34.5} \\
        TransRefer3D \cite{he2021TransRefer3DEA} & - & 57.4 & - & - & -\\
        SAT-2D \cite{Yang2021SAT2S}$^{*}$ & 35.4$^{\dagger}$ & \underline{57.9} & \underline{31.7}$^{\dagger}$ & 44.5 & 30.1\\
        \hline
        MDETR-\cite{Kamath2021MDETRM}-3D (our impl.) & 45.4 & - & 31.5 & 47.2 & 31.9 \\
        \hline
        \model (ours) & \textbf{52.1} & \textbf{67.0} & \textbf{43.3} & \textbf{52.2} & \textbf{39.8}\\
        \hline 
    \end{tabular}}
    \label{table:3d_results}
\end{table*}

We show quantitative results of our models against previous works 
in  Table~\ref{table:3d_results}. 
We use top-1 accuracy metric, which measures the percentage of times we can find the target box with an IoU higher than the threshold. We report results with IoU@0.25 on SR3D and NR3D; and with both IoU@0.25 and IoU@0.5 on ScanRefer. Please refer to supplementary for more detailed results.  

\begin{figure*}[t]
	\centering
	\includegraphics[width=\textwidth]{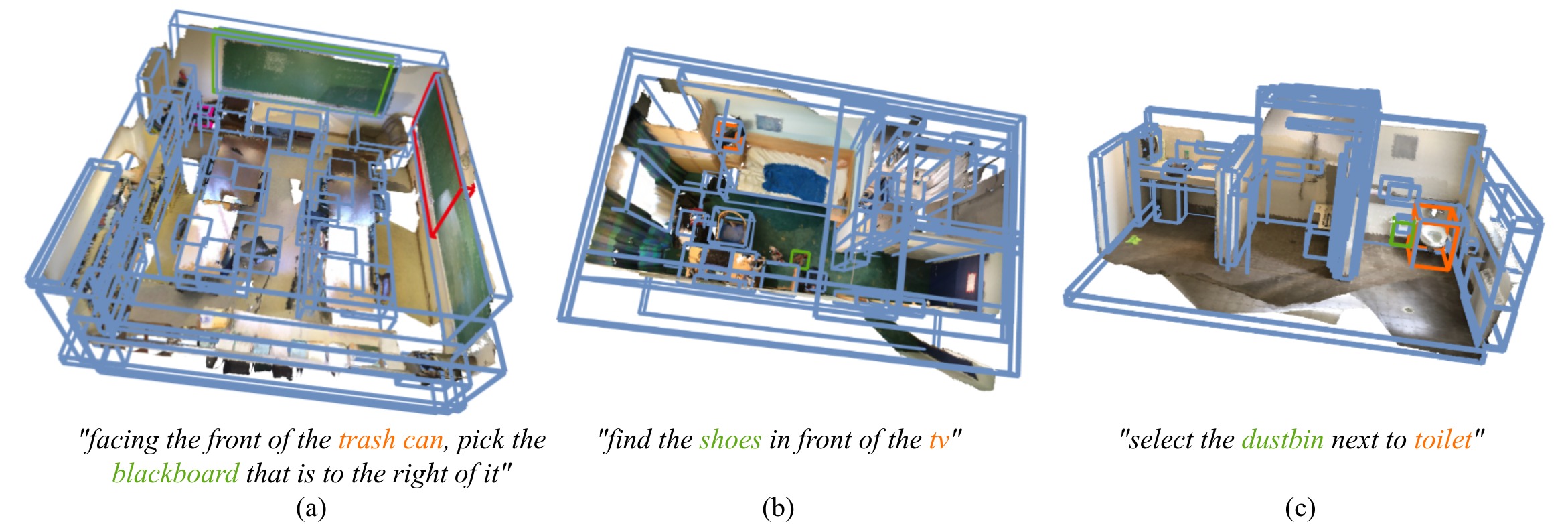}
	\caption{ 
	\textbf{Qualitative results of \model in the SR3D benchmark}. Predictions for the target are shown in green and for other mentioned objects in orange. The detected proposals appear in blue. (a) The variant without box stream (red box) fails to exploit the information given by the detector, but \model succeeds. (b) The detector misses the ``shoes'' and any box-bottlenecked variant fails. (c) The detector is successful in finding the ``dustbin'', still \model refines the box to get a more accurate bounding box.
	}
	\label{fig:qualitative}
\end{figure*}

\model outperforms existing  approaches as well as MDETR-3D by a large margin under both evaluation setups, $\Det$ and $\GT$. 
It also  outperforms the recent SAT-2D \cite{Yang2021SAT2S} that uses additional 2D RGB image features during training. $\model$ does not use 2D image features, but it can be easily extended to do so.   
We show qualitative results in Figure \ref{fig:qualitative}. For more qualitative results, please check the supplementary file. 


\begin{table}[t]
    \centering
    \caption{\textbf{Ablation of design choices for \model on SR3D}.}
    \resizebox{.98\textwidth}{!}
    {
    \begin{tabular}{|l|c|}
        \hline
        Model & Accuracy \\ \hline
        \model & \textbf{52.1} \\
        \quad w/o visual tokens & 41.9 \\ 
        \quad w/o detection prompts & 47.9 \\ 
        \quad w/o box stream & 51.0 \\ 
        \quad with MDETR's \cite{Kamath2021MDETRM} contrastive loss & 49.6 \\
        
        \quad w/o detection prompts; w/o box stream; (MDETR \cite{Kamath2021MDETRM}-3D) & 45.4 \\
        \quad with parametric queries; w/o detection prompts; w/o box stream; (MDETR \cite{Kamath2021MDETRM}-3D-Param) & 33.8 \\
        
        \quad with concatenated Visual, Language and Object Streams & 51.3 \\
        \hline
    \end{tabular}}
    \label{table:ablation}
\end{table}

\subsubsection{Ablative analysis}

We ablate all our design choices for 3D \model on SR3D benchmark \cite{Achlioptas2020ReferIt3DNL} in Table~\ref{table:ablation}. We compare 
\model against the following variants:
\begin{itemize}
    
    \item w/o visual tokens: an object-bottlenecked variant, which only attends to the language and box proposal streams and selects one box out of the proposals.
    
    \item w/o detection prompts: \model trained solely on SR3D grounding utterances. 
    
    \item w/o box stream: \model without attention on the box stream.
    
    \item w/ MDETR's contrastive loss: \model where we replace our modified contrastive loss with MDETR's.

    \item w/o detection prompts, w/o box stream, w/ MDETR's contrastive loss: an MDETR \cite{Kamath2021MDETRM}-3D implementation.
    
    \item w/ parametric queries, w/o detection prompts, w/o box stream, w/ MDETR's contrastive loss: an MDETR-3D implementation that uses  parametric object queries, as in original MDETR.
    
    \item w/ concatenated visual, language and box streams: instead of attending to each modality separately, we concatenate the different streams along their sequence dimension.
\end{itemize}

\paragraph{}The conclusions are as follows:

\begin{enumerate}
    \item \textbf{Box bottlenecks hurt:}  Models such as \model and MDETR-3D that decode object boxes instead of selecting them from a pool of given object proposals significantly outperform box-bottlenecked variants. 
    $\model$ outperforms by 10.2\% an object-bottlenecked variant, that does not attend to 3D point features and does not decode boxes.
    \item \textbf{$\model$  outperforms MDETR-3D by 6.7\%}: 
    \item \textbf{Attention on a box proposal stream helps:} Removing attention on the box stream  causes an absolute 1.1\% drop in accuracy. 
    \item \textbf{Co-training with detection prompts helps:} Co-training with detection prompts contributes 4.2\% in performance (from 47.9\% to 52.1\%).  
    \item \textbf{\model's contrastive loss helps:} Replacing our contrastive loss with MDETR's results in drop of 2.5\% in absolute accuracy. 
    \item \textbf{Concatenating Visual, Language and Object Streams performs worse than a model that has separate streams for each modality} Our motivation is to keep separate streams in 3D cross-modality encoder and decoder to be consistent with 2D \model as explained in Section \ref{sec:architecture}. We additionally find that having separate streams gives a boost of 0.8\%.

\end{enumerate}

\begin{table*}[t]
    \centering
    \caption{\textbf{Results on language grounding in 2D RefCOCO and RefCOCO+ Datasets on Top-1 accuracy metric using standard splits}. All training times are computed using same V100 GPUs. Training epochs are written as $x+y$ where $x$ = number of pre-training epochs and $y$ = number of fine-tuning epochs. All reported results use ResNet101 backbone.}
    {
    \begin{tabular}{|l||*{9}{c|}}
        \hline
        & \multicolumn{3}{c|}{RefCOCO} & \multicolumn{3}{c|}{RefCOCO+} & Training & Training
        \\
        Method & val & testA & testB & val & testA & testB & Epochs & GPU Hours \\ \hline
        UNITER\_L \cite{Chen2020UNITERUI} &  81.4 & 87.0 & 74.2 & 75.9 & 81.5 & 66.7 & - & - \\
        VILLA\_L \cite{Gan2020LargeScaleAT} &  82.4 & 87.5 & 74.8 & 76.2 & 81.5 & 66.8 & - & - \\ 
        MDETR \cite{Kamath2021MDETRM} & \textbf{86.8} & \textbf{89.6} & 81.4 & \textbf{79.5} & \textbf{84.1} & \textbf{70.6} &  40 + 5 & 5560 \\ \hline
        \model (ours) & 85.9 & 88.5 & \textbf{81.5} & 78.2 & 82.8 & 70.0 &  \textbf{12 + 5} & \textbf{2748} \\ \hline
    \end{tabular}}
    \label{table:2d_results_refcoco}
\end{table*}

\begin{table*}
    \centering
    \caption{\textbf{Results on language grounding in Flickr30k 2D images.} We use Recall@k metric. All training times are computed using same V100 GPUs.}
    {
    \begin{tabular}{|l||*{8}{c|}}
        \hline
        & \multicolumn{3}{c|}{Val} & \multicolumn{3}{c|}{Test} & Training & Training\\
        Method & R@1 & R@5 & R@10 & R@1 & R@5 & R@10 & Epochs & GPU hours \\ \hline
        VisualBERT \cite{Li2019VisualBERTAS} & 70.4 & 84.5 & 86.3 & 71.3 & 85.0 & 86.5 & - & - \\
        MDETR \cite{Kamath2021MDETRM} & \textbf{82.5} & \textbf{92.9} & \textbf{94.9} & \textbf{83.4} & \textbf{93.5} & \textbf{95.3} & 40 & 5480 \\ \hline
        \model (ours) & 81.2 & 90.9  & 92.8 & 81.0 & 91.6 & 93.2 & \textbf{12} & \textbf{2688} \\ \hline
    \end{tabular}}
    \label{table:2d_results_flickr}
\end{table*}

\subsection{Language grounding in 2D images}

We test \model on the referential grounding  datasets of RefCOCO \cite{Kazemzadeh2014ReferItGameRT}, RefCOCO+ \cite{Yu2016ModelingCI} and Flickr30k entities dataset \cite{Plummer2015Flickr30kEC}. 
We follow the pretrain-then-finetune  protocol of MDETR and  first pre-train on combined grounding annotations from Flickr30k \cite{Plummer2015Flickr30kEC}, referring expression datasets \cite{Kazemzadeh2014ReferItGameRT,Yu2016ModelingCI,Mao2016GenerationAC}, Visual Genome \cite{Krishna2016VisualGC}. During pre-training the task is to detect all instances of objects mentioned in the utterance. Different than MDETR, we augment this supervision with  detection prompts from the MS-COCO dataset \cite{Lin2014MicrosoftCC}. Following MDETR, we directly evaluate our pre-trained model on Flickr30k without any further fine-tuning and fine-tune for 5 epochs on RefCOCO and RefCOCO+.

\begin{table}[t]
    \centering
    \caption{\textbf{Ablation for \model on the RefCOCO validation set}.}
    \resizebox{.95\textwidth}{!}
    {
    \begin{tabular}{|p{0.95\textwidth}|c|}
        \hline
        Model & Accuracy \\
        \hline 
        \model & \textbf{79.4} \\
        \quad w/o det prompts & 77.0 \\
        \quad w/o box stream w/o det prompts & 76.3 \\ 
        \quad  w/o box stream w/o det prompts w/ MDETR's \cite{Kamath2021MDETRM} contrastive & 74.2 \\ \hline
    \end{tabular}}
    \label{table:ablation_2d}
\end{table}

We report top-1 accuracy on the standard splits of RefCOCO and RefCOCO+  in Table~\ref{table:2d_results_refcoco} and Recall metric with ANY-BOX protocol \cite{Li2019VisualBERTAS} on Flickr30k in Table~\ref{table:2d_results_flickr}. Our model and MDETR use the same 200k image-language pairs from COCO \cite{Lin2014MicrosoftCC}, Flickr30k \cite{Plummer2015Flickr30kEC} and Visual Genome \cite{Krishna2016VisualGC}. VisualBERT \cite{Li2019VisualBERTAS} is trained on COCO captions. UNITER \cite{Chen2020UNITERUI} and VILLA \cite{Gan2020LargeScaleAT} use a larger dataset of 4.4M pairs from COCO, Visual Genome,  Conceptual-Captions \cite{sharma2018conceptual}, and SBU Captions \cite{ordonez2011im2text}. In addition, we augment our training set with detection prompts from COCO.
$\model$ trains two times faster than MDETR while getting comparable performance. This computational gain comes mostly from deformable attention which is much cheaper than original visual self-attention that scales quadratically with the number of visual tokens, as already reported in \cite{Zhu2021DeformableDD}. 
For qualitative results, please see the supplementary file.

\subsubsection{Ablative analysis}

We ablate our model in RefCOCO without pre-training in Table~\ref{table:ablation_2d}, since pre-training is computationally expensive due to the size of the combined datasets. Consistent with 3D, removing detection prompts results in an accuracy drop of 2.4\%. Additionally removing attention to the box proposal stream results in a drop of 3.1\% in accuracy. When replacing our contrastive loss with MDETR's, the model achieves 74.2\%, resulting in an additional drop of 2.1\% accuracy. 

\subsection{Limitations}
Our work relies on language-image alignment and does not address how to ground  language 
 better and more robustly through abstraction of the visual features, e.g., the fact that \textit{left} and \textit{right} reverse when we change the user's viewpoint, the fact that numbers requires precise counting, or the fact that the \textit{`` chair furthest away from the door"} requires to satisfy a logical constraint  which our model can totally violate when presented with out-of-distribution visual input. 
This limitation is a  direct avenue for future work.



\section{Conclusion}
We present $\model$, a model for referential grounding in 3D and 2D scenes, that attends to  language, visual and box proposal  streams to decode objects mentioned in the referential utterance and align them to corresponding spans in the input. 
$\model$ builds upon   MDETR \cite{Kamath2021MDETRM} and outperforms  its straightforward MDETR-3D equivalent by a significant margin thanks to attention on labelled bottom-up box proposals, co-training with detection prompts and improved contrastive losses, setting a new state-of-the-art in two 3D language grounding benchmarks.  
\model is also the first model in 3D referential grounding that operates on the realistic setup of not having access to oracle object boxes, but rather detects them from the input 3D point cloud. 
\newline




\noindent \textbf{Acknowledgement} This material is based upon work supported by an Air Force grant FA95502010423, an AFOSR Young Investigator Award,  DARPA Machine Common Sense, and an NSF CAREER award provided in affiliation to Carnegie Mellon University.  Any opinions, findings and conclusions or recommendations expressed in this material are those of the author(s) and do not necessarily reflect the views of the United States Army or the United States Air Force. The authors would also like to thank Adam W. Harley, Leonid Keselman, Muyang Li and Rohan Choudhary for their helpful feedback.

{\small
\bibliographystyle{splncs04}
\bibliography{egbib,lang,iclr2022_conference}

\begin{thebibliography}{10}
\providecommand{\url}[1]{\texttt{#1}}
\providecommand{\urlprefix}{URL }
\providecommand{\doi}[1]{https://doi.org/#1}

\bibitem{Abdelreheem20223DRefTransformerFO}
Abdelreheem, A., Upadhyay, U., Skorokhodov, I., Yahya, R.A., Chen, J.,
  Elhoseiny, M.: {3DRefTransformer: Fine-Grained Object Identification in
  Real-World Scenes Using Natural Language}. In: Proc. WACV (2022)

\bibitem{Achlioptas2020ReferIt3DNL}
Achlioptas, P., Abdelreheem, A., Xia, F., Elhoseiny, M., Guibas, L.:
  {ReferIt3D: Neural Listeners for Fine-Grained 3D Object Identification in
  Real-World Scenes}. In: Proc. ECCV (2020)

\bibitem{Anderson2018BottomUpAT}
Anderson, P., He, X., Buehler, C., Teney, D., Johnson, M., Gould, S., Zhang,
  L.: {Bottom-Up and Top-Down Attention for Image Captioning and Visual
  Question Answering}. In: Proc. CVPR (2018)

\bibitem{Carion2020EndtoEndOD}
Carion, N., Massa, F., Synnaeve, G., Usunier, N., Kirillov, A., Zagoruyko, S.:
  {End-to-End Object Detection with Transformers}. In: Proc. ECCV (2020)

\bibitem{Chen2020ScanRefer3O}
Chen, D.Z., Chang, A., Nie{\ss}ner, M.: {ScanRefer: 3D Object Localization in
  RGB-D Scans using Natural Language}. In: Proc. ECCV (2020)

\bibitem{chen2018real}
Chen, X., Ma, L., Chen, J., Jie, Z., Liu, W., Luo, J.: Real-time referring
  expression comprehension by single-stage grounding network. ArXiv
  \textbf{abs/1812.03426} (2018)

\bibitem{Chen2020UNITERUI}
Chen, Y.C., Li, L., Yu, L., Kholy, A.E., Ahmed, F., Gan, Z., Cheng, Y., Liu,
  J.: {UNITER: UNiversal Image-TExt Representation Learning}. In: Proc. ECCV
  (2020)

\bibitem{Dai2017ScanNetR3}
Dai, A., Chang, A.X., Savva, M., Halber, M., Funkhouser, T.A., Nie{\ss}ner, M.:
  {ScanNet: Richly-Annotated 3D Reconstructions of Indoor Scenes}. In: Proc.
  CVPR (2017)

\bibitem{Deng2009ImageNetAL}
Deng, J., Dong, W., Socher, R., Li, L.J., Li, K., Fei-Fei, L.: {ImageNet: A
  large-scale hierarchical image database}. In: Proc. CVPR (2009)

\bibitem{deng2021transvg}
Deng, J., Yang, Z., Chen, T., Zhou, W., Li, H.: Transvg: End-to-end visual
  grounding with transformers. In: Proc. ICCV (2021)

\bibitem{fang2015captions}
Fang, H., Gupta, S., Iandola, F., Srivastava, R.K., Deng, L., Doll{\'a}r, P.,
  Gao, J., He, X., Mitchell, M., Platt, J.C., et~al.: {From Captions to Visual
  Concepts and Back}. In: Proc. CVPR (2015)

\bibitem{Feng2021FreeformDG}
Feng, M., Li, Z., Li, Q., Zhang, L., Zhang, X., Zhu, G., Zhang, H., Wang, Y.,
  Mian, A.: {Free-form Description Guided 3D Visual Graph Network for Object
  Grounding in Point Cloud}. In: Proc. ICCV (2021)

\bibitem{Fukui2016MultimodalCB}
Fukui, A., Park, D.H., Yang, D., Rohrbach, A., Darrell, T., Rohrbach, M.:
  {Multimodal Compact Bilinear Pooling for Visual Question Answering and Visual
  Grounding}. In: Proc. EMNLP (2016)

\bibitem{Gan2020LargeScaleAT}
Gan, Z., Chen, Y.C., Li, L., Zhu, C., Cheng, Y., Liu, J.: {Large-Scale
  Adversarial Training for Vision-and-Language Representation Learning}. In:
  Proc. NeurIPS (2020)

\bibitem{he2021TransRefer3DEA}
He, D., Zhao, Y., Luo, J., Hui, T., Huang, S., Zhang, A., Liu, S.:
  {TransRefer3D: Entity-and-Relation Aware Transformer for Fine-Grained 3D
  Visual Grounding}. In: Proc. ACMMM (2021)

\bibitem{DBLP:journals/corr/HeGDG17}
He, K., Gkioxari, G., Doll{\'{a}}r, P., Girshick, R.B.: Mask {R-CNN}. In: Proc.
  ICCV (2017)

\bibitem{he2016deep}
He, K., Zhang, X., Ren, S., Sun, J.: Deep residual learning for image
  recognition. In: Proceedings of the IEEE Conference on Computer Vision and
  Pattern Recognition (CVPR). pp. 770--778 (2016)

\bibitem{He2016DeepRL}
He, K., Zhang, X., Ren, S., Sun, J.: {Deep Residual Learning for Image
  Recognition}. In: Proc. CVPR (2016)

\bibitem{modularreferential}
Hu, R., Rohrbach, M., Andreas, J., Darrell, T., Saenko, K.: {Modeling
  Relationships in Referential Expressions with Compositional Modular
  Networks}. In: Proc. CVPR (2017)

\bibitem{Huang2021TextGuidedGN}
Huang, P.H., Lee, H.H., Chen, H.T., Liu, T.L.: {Text-Guided Graph Neural
  Networks for Referring 3D Instance Segmentation}. In: Proc. AAAI (2021)

\bibitem{Jaegle2021PerceiverGP}
Jaegle, A., Gimeno, F., Brock, A., Zisserman, A., Vinyals, O., Carreira, J.:
  {Perceiver: General Perception with Iterative Attention}. In: Proc. ICML
  (2021)

\bibitem{densecap}
Johnson, J., Karpathy, A., Fei-Fei, L.: {DenseCap: Fully Convolutional
  Localization Networks for Dense Captioning}. In: Proc. CVPR (2016)

\bibitem{Kamath2021MDETRM}
Kamath, A., Singh, M., LeCun, Y.A., Misra, I., Synnaeve, G., Carion, N.: {MDETR
  - Modulated Detection for End-to-End Multi-Modal Understanding}. In: Proc.
  ICCV (2021)

\bibitem{karpathy2015deep2}
Karpathy, A., Fei-Fei, L.: {Deep Visual-semantic Alignments for Generating
  Image Descriptions}. In: Proc. CVPR (2015)

\bibitem{Kazemzadeh2014ReferItGameRT}
Kazemzadeh, S., Ordonez, V., Matten, M.A., Berg, T.L.: {ReferItGame: Referring
  to Objects in Photographs of Natural Scenes}. In: Proc. EMNLP (2014)

\bibitem{Krishna2016VisualGC}
Krishna, R., Zhu, Y., Groth, O., Johnson, J., Hata, K., Kravitz, J., Chen, S.,
  Kalantidis, Y., Li, L.J., Shamma, D.A., Bernstein, M.S., Fei-Fei, L.: {Visual
  Genome: Connecting Language and Vision Using Crowdsourced Dense Image
  Annotations}. International Journal of Computer Vision  \textbf{123} (2016)

\bibitem{Li2019VisualBERTAS}
Li, L.H., Yatskar, M., Yin, D., Hsieh, C.J., Chang, K.W.: {VisualBERT: A Simple
  and Performant Baseline for Vision and Language}. ArXiv
  \textbf{abs/1908.03557} (2019)

\bibitem{Li2021GroundedLP}
Li, L.H., Zhang, P., Zhang, H., Yang, J., Li, C., Zhong, Y., Wang, L., Yuan,
  L., Zhang, L., Hwang, J.N., Chang, K.W., Gao, J.: {Grounded Language-Image
  Pre-training}. In: Proc. CVPR (2022)

\bibitem{Lin2017FocalLF}
Lin, T.Y., Goyal, P., Girshick, R.B., He, K., Doll{\'a}r, P.: {Focal Loss for
  Dense Object Detection}. In: Proc. ICCV (2017)

\bibitem{Lin2014MicrosoftCC}
Lin, T.Y., Maire, M., Belongie, S.J., Hays, J., Perona, P., Ramanan, D.,
  Doll{\'a}r, P., Zitnick, C.L.: {Microsoft COCO: Common Objects in Context}.
  In: Proc. ECCV (2014)

\bibitem{DBLP:journals/corr/LiuAESR15}
Liu, W., Anguelov, D., Erhan, D., Szegedy, C., Reed, S.E., Fu, C., Berg, A.C.:
  {SSD: Single Shot MultiBox Detector}. In: Proc. ECCV (2016)

\bibitem{Liu2019RoBERTaAR}
Liu, Y., Ott, M., Goyal, N., Du, J., Joshi, M., Chen, D., Levy, O., Lewis, M.,
  Zettlemoyer, L., Stoyanov, V.: {RoBERTa: A Robustly Optimized BERT
  Pretraining Approach}. ArXiv  \textbf{abs/1907.11692} (2019)

\bibitem{Liu2021GroupFree3O}
Liu, Z., Zhang, Z., Cao, Y., Hu, H., Tong, X.: {Group-Free 3D Object Detection
  via Transformers}. In: Proc. ICCV (2021)

\bibitem{Lu2019ViLBERTPT}
Lu, J., Batra, D., Parikh, D., Lee, S.: {ViLBERT: Pretraining Task-Agnostic
  Visiolinguistic Representations for Vision-and-Language Tasks}. In: Proc.
  NeurIPS (2019)

\bibitem{Lu202012in1MV}
Lu, J., Goswami, V., Rohrbach, M., Parikh, D., Lee, S.: {12-in-1: Multi-Task
  Vision and Language Representation Learning}. In: Proc. CVPR (2020)

\bibitem{Mao2016GenerationAC}
Mao, J., Huang, J., Toshev, A., Camburu, O.M., Yuille, A.L., Murphy, K.P.:
  {Generation and Comprehension of Unambiguous Object Descriptions}. In: Proc.
  CVPR (2016)

\bibitem{Misra2021AnET}
Misra, I., Girdhar, R., Joulin, A.: {An End-to-End Transformer Model for 3D
  Object Detection}. In: Proc. ICCV (2021)

\bibitem{ordonez2011im2text}
Ordonez, V., Kulkarni, G., Berg, T.: {Im2text: Describing images using 1
  million captioned photographs}. In: Proc. NIPS (2011)

\bibitem{Plummer2015Flickr30kEC}
Plummer, B.A., Wang, L., Cervantes, C.M., Caicedo, J.C., Hockenmaier, J.,
  Lazebnik, S.: {Flickr30k Entities: Collecting Region-to-Phrase
  Correspondences for Richer Image-to-Sentence Models}. In: Proc. ICCV (2015)

\bibitem{Qi2019DeepHV}
Qi, C., Litany, O., He, K., Guibas, L.J.: {Deep Hough Voting for 3D Object
  Detection in Point Clouds}. In: Proc. ICCV (2019)

\bibitem{Qi2017PointNetDH}
Qi, C., Yi, L., Su, H., Guibas, L.J.: {PointNet++: Deep Hierarchical Feature
  Learning on Point Sets in a Metric Space}. In: Proc. NIPS (2017)

\bibitem{Redmon2016YouOL}
Redmon, J., Divvala, S.K., Girshick, R.B., Farhadi, A.: {You Only Look Once:
  Unified, Real-Time Object Detection}. In: Proc. CVPR (2016)

\bibitem{ren2015faster}
Ren, S., He, K., Girshick, R., Sun, J.: {Faster R-CNN: Towards Real-time Object
  Detection with Region Proposal Networks}. In: Proc. NIPS (2015)

\bibitem{Rezatofighi2019GeneralizedIO}
Rezatofighi, S.H., Tsoi, N., Gwak, J., Sadeghian, A., Reid, I.D., Savarese, S.:
  {Generalized Intersection Over Union: A Metric and a Loss for Bounding Box
  Regression}. In: Proc. CVPR (2019)

\bibitem{Roh2021LanguageReferSM}
Roh, J., Desingh, K., Farhadi, A., Fox, D.: {LanguageRefer: Spatial-Language
  Model for 3D Visual Grounding}. In: Proc. CoRL (2021)

\bibitem{sadhu2019zero}
Sadhu, A., Chen, K., Nevatia, R.: Zero-shot grounding of objects from natural
  language queries. In: Pro. ICCV (2019)

\bibitem{sharma2018conceptual}
Sharma, P., Ding, N., Goodman, S., Soricut, R.: {Conceptual Captions: A
  Cleaned, Hypernymed, Image Alt-text Dataset for Automatic Image Captioning}.
  In: Proc. ACL (2018)

\bibitem{Vaswani2017AttentionIA}
Vaswani, A., Shazeer, N.M., Parmar, N., Uszkoreit, J., Jones, L., Gomez, A.N.,
  Kaiser, L., Polosukhin, I.: {Attention is All you Need}. In: Proc. NIPS
  (2017)

\bibitem{yang2020improving}
Yang, Z., Chen, T., Wang, L., Luo, J.: {Improving one-stage visual grounding by
  recursive sub-query construction}. In: Proc. ECCV (2020)

\bibitem{yang2019fast}
Yang, Z., Gong, B., Wang, L., Huang, W., Yu, D., Luo, J.: A fast and accurate
  one-stage approach to visual grounding. In: Proc. ICCV (2019)

\bibitem{Yang2021SAT2S}
Yang, Z., Zhang, S., Wang, L., Luo, J.: {SAT: 2D Semantics Assisted Training
  for 3D Visual Grounding}. In: Proc. ICCV (2021)

\bibitem{Yu2016ModelingCI}
Yu, L., Poirson, P., Yang, S., Berg, A.C., Berg, T.L.: {Modeling Context in
  Referring Expressions}. In: Proc. ECCV (2016)

\bibitem{Yu2018RethinkingDA}
Yu, Z., Yu, J., Xiang, C., Zhao, Z., Tian, Q., Tao, D.: {Rethinking Diversified
  and Discriminative Proposal Generation for Visual Grounding}. In: Proc. IJCAI
  (2018)

\bibitem{Yuan2021InstanceReferCH}
Yuan, Z., Yan, X., Liao, Y., Zhang, R., Li, Z., Cui, S.: {InstanceRefer:
  Cooperative Holistic Understanding for Visual Grounding on Point Clouds
  through Instance Multi-level Contextual Referring}. In: Proc. ICCV (2021)

\bibitem{Zhao_2021_ICCV}
Zhao, L., Cai, D., Sheng, L., Xu, D.: {3DVG-Transformer: Relation Modeling for
  Visual Grounding on Point Clouds}. In: Proc. ICCV (2021)

\bibitem{Zhu2021DeformableDD}
Zhu, X., Su, W., Lu, L., Li, B., Wang, X., Dai, J.: {Deformable DETR:
  Deformable Transformers for End-to-End Object Detection}. In: Proc. ICLR
  (2021)

\end{thebibliography}
}
\newpage

\section{Supplementary file}
\subsection{Overview}
In Section~\ref{implement}, we provide implementation details for \model on both the 3D and the 2D domain. In Section~\ref{moreResults}, we provide a detailed analysis of our results on SR3D, NR3D \cite{Achlioptas2020ReferIt3DNL} and ScanRefer benchmarks \cite{Chen2020ScanRefer3O}. In Section~\ref{det_backbone} we ablate the choice of the detection backbone and experiment with unfreezing it during the referential grounding training stage. In Section~\ref{det_augment}, we show the effect of corrupting the detector's proposals at training time. In Section~\ref{negatives}, we discuss training with detection prompts that contain negative labels. We evaluate our model as a language-modulated object detector in Section~\ref{det_results}. In Section~\ref{qual}, we show more qualitative results on both 3D point clouds and 2D images, including failure cases.


\subsection{Implementation details}
\label{implement}
We report here architecture choices as well as training hyperparameters. We implement \model in PyTorch.
For the 3D version, the point cloud is encoded with PointNet++ \cite{Qi2017PointNetDH} using the same hyperparameters as in \cite{Liu2021GroupFree3O}, pre-trained on ScanNet \cite{Dai2017ScanNetR3}. We use the last layer's features, resulting in 1024 visual tokens. The detected boxes are encoded using their spatial and categorical features. Specifically, we encode each box's coordinates with an MLP, then we concatenate this vector with projected RoBERTa \cite{Liu2019RoBERTaAR} embeddings and feed to another MLP to obtain the box embeddings. For the cross-modality encoder, we use $N_E=3$ layers. All attention layers are implemented using standard key-value attention \cite{Vaswani2017AttentionIA,Lu2019ViLBERTPT}. 
In the decoder, the queries are formed from the 256 most confident visual tokens. To compute this confidence score, each visual token is fed to an MLP to give a scalar value. We supervise these values using Focal Loss \cite{Lin2017FocalLF}. Specifically, since each visual token corresponds to a point with known coordinates, we associate visual tokens to ground-truth object centers and keep the 4 closest points to each center. We consider these matched points as positives, i.e. here points with high ground-truth objectness. The same scoring method is employed in \cite{Liu2021GroupFree3O}. We use $N_D=6$ decoder layers. Similar to encoder, all attention layers are implemented using standard self-/cross-attention.

For the 2D version, the image is encoded using ResNet-101 \cite{He2016DeepRL} pretrained on ImageNet \cite{Deng2009ImageNetAL}. We  use multi-scale features as in \cite{Zhu2021DeformableDD}. The feature maps of the different scales are flattened and concatenated in the spatial dimension, leading to 17821 visual tokens. The feature dimension of each token is 256. 
To obtain the box proposals, we use the detector of \cite{Anderson2018BottomUpAT} trained on 1601 classes of Visual Genome \cite{Krishna2016VisualGC}. The detected boxes are encoded using their spatial and categorical features. Specifically, we compute the 2D Fourier features of each box and feed them to an MLP, then we concatenate this vector with projected RoBERTa \cite{Liu2019RoBERTaAR} embeddings and feed to another MLP to obtain the box embeddings. 
To form queries, we rank visual tokens based on their confidence score and keep the 300 most confidence ones. This confidence layer is supervised using Focal Loss \cite{Lin2017FocalLF}: we assign a positive objectness scores to every point that lies inside a ground-truth answer box. We set $N_E=6$ and $N_D=6$. All attention layers to the visual stream are implemented with deformable attention \cite{Zhu2021DeformableDD}, attention to either the language stream or detected boxes is the standard attention of \cite{Vaswani2017AttentionIA,Lu2019ViLBERTPT}. We do not use deformable attention in the 3D domain since computing it requires pooling features and doing bilinear interpolation from neighbouring pixels. In 2D, finding neighbouring pixels can be trivially done by simply looking up neighbouring indices due to its continuous grid structure. However, in discontinous domains like 3D, we would need to compute all pairs of distances between the points in a given pointcloud and rank them to obtain nearest neighbours. This is computationally expensive. Moreover, since pointclouds have irregular density, using a fixed number of neighbours is sub-optimal. These issues can be resolved by using specialised data-structures like KD-Trees and by using adaptive neighbourhood sampling, however they are beyond the scope of this work.

\begin{table}[t]
    \centering
    \caption{\textbf{Performance analysis on language grounding on SR3D.} We evaluate top-1 accuracy using ground-truth ($\GT$) boxes, under the different setups introduced in \cite{Achlioptas2020ReferIt3DNL}. See the main text for an explanation of each setup.}
    {
    \begin{tabular}{|l||*{5}{c|}}
        \hline
        Method & Easy & Hard & View-Dep & View-Indep & Overall ($\GT$) \\
        \hline
        ReferIt3DNet \cite{Achlioptas2020ReferIt3DNL} & 44.7 & 31.5 & 39.2 & 40.8 & 39.8\\
        TGNN \cite{Huang2021TextGuidedGN} & 48.5 & 36.9 & 45.8 & 45.0 & 45.0 \\
        3DRefTransformer \cite{Abdelreheem20223DRefTransformerFO} & 50.7 & 38.3 & 44.3 & 47.1 & 47.0 \\
        InstanceRefer \cite{Yuan2021InstanceReferCH} & 51.1 & 40.5 & 45.4 & 48.1 & 48.0 \\
        LanguageRefer \cite{Roh2021LanguageReferSM} & 58.9 & 49.3 & 49.2 & 56.3 & 56.0 \\
        3DVG-Transformer \cite{Zhao_2021_ICCV} & 54.2 & 44.9 & 44.6 & 51.7 & 51.4 \\
        TransRefer3D \cite{he2021TransRefer3DEA} & 60.5 & \underline{50.2} & \underline{49.9} & 57.7 & 57.4 \\
        SAT 2D \cite{Yang2021SAT2S} & \underline{61.2} & 50.0 & 49.2 & \underline{58.3} & \underline{57.9}\\
        \hline
        \model (ours) & \textbf{68.6} & \textbf{63.2} & \textbf{53.0} & \textbf{67.6} & \textbf{67.0} \\
        \hline
    \end{tabular}}
    \label{table:sr3d_results}
\end{table}

\begin{table}[t]
    \centering
    \caption{\textbf{Performance analysis on language grounding on NR3D.} We evaluate top-1 accuracy using ground-truth ($\GT$) boxes, under the different setups introduced in \cite{Achlioptas2020ReferIt3DNL}. See the main text for an explanation of each setup.}
    {
    \begin{tabular}{|l||*{5}{c|}}
        \hline
        Method & Easy & Hard & View-Dep & View-Indep & Overall (GT) \\
        \hline
        ReferIt3DNet \cite{Achlioptas2020ReferIt3DNL} & 43.6 & 27.9 & 32.5 & 37.1 & 35.6\\
        TGNN \cite{Huang2021TextGuidedGN} & 44.2 & 30.6 & 35.8 & 38.0 & 37.3 \\
        3DRefTransformer \cite{Abdelreheem20223DRefTransformerFO} & 46.4 & 32.0 & 34.7 & 41.2 & 39.0 \\
        InstanceRefer \cite{Yuan2021InstanceReferCH} & 46.0 & 31.8 & 34.5 & 41.9 & 38.8 \\
        FFL-3DOG \cite{Feng2021FreeformDG} & 48.2 & 35.0 & 37.1 & 44.7 & 41.7 \\
        LanguageRefer \cite{Roh2021LanguageReferSM} & 51.0 & 36.6 & 41.7 & 45.0 & 43.9 \\
        3DVG-Transformer \cite{Zhao_2021_ICCV} & 48.5 & 34.8 & 34.8 & 43.7 & 40.8 \\
        TransRefer3D \cite{he2021TransRefer3DEA} & 48.5 & 36.0 & 36.5 & 44.9 & 42.1 \\
        SAT 2D \cite{Yang2021SAT2S} & \underline{56.3} & \underline{42.4} & \textbf{46.9} & \underline{50.4} & \underline{49.2} \\
        \hline
        \model (ours) & \textbf{60.7} & \textbf{48.4} & \underline{46.0} & \textbf{58.0} & \textbf{54.6} \\
        \hline
    \end{tabular}}
    \label{table:nr3d_results}
\end{table}

For the 3D model, we freeze the text encoder and use a learning rate of $1e{-3}$ for the visual encoder and $1e{-4}$ for all other layers. We are able to fit a batch size of 6 on a single GPU of 12GB and 24 on an NVIDIA A100. Under these conditions, each epoch takes around 50 minutes on an A100. For the 2D model, we use a learning rate of $1e{-6}$ for Resnet101 visual encoder, $5e{-6}$ for RoBERTa text encoder and $1e{-5}$ for rest of the layers. We pre-train on 64 NVIDIA V100 GPUs with a batch size of 1, and finetune on RefCOCO/RefCOCO+ with a batch size of 2 on 16 V100s. The total training time is included in the respective tables. We release pre-trained checkpoints for both 3D and 2D models.

\begin{table}
    \centering
    \caption{\textbf{Performance analysis on language grounding on ScanRefer.} We evaluate top-1 accuracy using detected boxes, under the different setups introduced in \cite{Chen2020ScanRefer3O}. See the main text for an explanation of each setup.}
    \resizebox{\textwidth}{!}
    {
    \begin{tabular}{|l||*{5}{c|}c|}
        \hline
        Method & Unique@0.25 & Unique@0.5 & Multi@0.25 & Multi@0.5 & Overall@0.25 & Overall@0.5 \\
        \hline
        ReferIt3DNet \cite{Achlioptas2020ReferIt3DNL} & 53.8 & 37.5 & 21.0 & 12.8 & 26.4 & 16.9\\
        ScanRefer \cite{Chen2020ScanRefer3O} & 63.0 & 40.0 & 28.9 & 18.2 & 35.5 & 22.4\\
        TGNN \cite{Huang2021TextGuidedGN} & 68.6 & 56.8 & 29.8 & 23.2 & 37.4 & 29.7 \\
        InstanceRefer \cite{Yuan2021InstanceReferCH} & 77.5 & \underline{66.8} & 31.3 & 24.8 & 40.2 & 32.9 \\
        FFL-3DOG \cite{Feng2021FreeformDG} & \underline{78.8} & \textbf{67.9} & 35.2 & 25.7 & 41.3 & 34.0 \\
        3DVG-Transformer \cite{Zhao_2021_ICCV} & 77.2 & 58.5 & \underline{38.4} & \underline{28.7} & \underline{45.9} & \underline{34.5} \\
        SAT 2D \cite{Yang2021SAT2S} & - & - & - & - & 44.5 & 30.1\\
        \hline
        \model (ours) & \textbf{84.2} & 66.3 & \textbf{46.6} & \textbf{35.1} & \textbf{52.2} & \textbf{39.8} \\
        \hline
    \end{tabular}}
    \label{table:scanrefer_results}
\end{table}

\subsection{Detailed results on SR3D/NR3D and ScanRefer} \label{moreResults}
We include results on SR3D/NR3D \cite{Achlioptas2020ReferIt3DNL} and ScanRefer \cite{Chen2020ScanRefer3O} under the different evaluation protocols specified in the original papers. Similar to prior works, we report results using overall accuracy metric. In $\det$ setup, we threshold over the IoU between the box regressed by \model and the ground truth box. In $\GT$ setup, we select the ground truth box that the has highest IoU with the most confident box regressed by \model and check if it matches with the target box. Besides overall accuracy, we additionally report accuracy on the following contexts for SR3D/NR3D:
\begin{itemize}
    \item Easy: there is only one ``distractor'', i.e. object belonging to the same class as the target instance
    \item Hard: there are two or more distractors
    \item View-dependent: cases for which rotating the scene around the z axis would lead to a different answer, e.g. ``tv left of sofa''
    \item View-independent: rotation does not affect the answer, e.g. ``chair closest to table''
\end{itemize}

We evaluate on the following contexts for ScanRefer:
\begin{itemize}
    \item Unique: there is no ``distractor'', i.e. object belonging to the same class as the target instance
    \item Multi: there is at least one distractor
\end{itemize}

We compare \model against prior approaches in Table~\ref{table:sr3d_results} for SR3D, Table~\ref{table:nr3d_results} for NR3D and Table~\ref{table:scanrefer_results} for ScanRefer. For SR3D and NR3D, all models are trained and tested with access to ground-truth object proposals, as in \cite{Achlioptas2020ReferIt3DNL}. For ScanRefer, all models are trained and tested with detected objects, so we report accuracy under the 0.25 and 0.5 IoU thresholds. We vastly outperform all competitors under all setups on SR3D. On NR3D, we show clear gains on all protocols except for view-dependent. Performance on this setup could be improved by incorporating a view prediction network, but we aimed to have a model that works for both 3D and 2D with as least domain-specific design choices as possible. On ScanRefer, we clearly outperform all previous approaches under all setups except for Unique@0.5, where we perform on par with the best-performing competitor.


\subsection{Effect of detection backbone}\label{det_backbone}
To examine the importance of the detection backbone, since previous work use VoteNet \cite{Qi2019DeepHV} as their detector, we evaluate our model using VoteNet boxes on ScanRefer and get 50.0\% Acc@0.25 and 37.5\% Acc@0.5 (in comparison to 50.9\% and 38.8\% with Group-Free boxes), which still outperforms all competitors. On SR3D and NR3D all previous works use GT boxes; hence we re-run all baselines of Table 1 with the same detector as our model.

Additionally, we try to unfreeze the object detector backbone during training with language. Inspired by \cite{Yu2018RethinkingDA}, we added a box regression layer in our baseline ``w/o visual tokens" of Table 2. This achieves 46.4\% on SR3D, which is indeed better than our previous
baseline by 4.5\%. However, it still underperforms our proposed model by 4.7\%. This result indicates that box-bottlenecked baselines still underperform, even when the object detector is finetuned.

\subsection{Effect of detection augmentation} \label{det_augment}
As we mention in the main paper, the 3D detector is trained on ScanNet and thus the proposals are of much better quality at train time and worse at test time. To mitigate overfitting, we randomly replace 30\% of the detected boxes at training time with random ones. Quantitatively, this gives a boost of 1\% absolute, as seen in Table~\ref{table:det_aug}. Note that this augmentation can only be applied when the box stream is employed.

\begin{table}[t]
    \centering
    \caption{\textbf{Effect of detection augmentation on SR3D.}}
    {
    \begin{tabular}{|l|c|}
        \hline
        Method & Overall (Det) \\
        \hline
        \model w/o box stream & 51.0\\
        \model w/o detection augmentation & 51.1 \\
        \model & 52.1 \\
        \hline
    \end{tabular}}
    \label{table:det_aug}
\end{table}

\subsection{Negative training with detection prompts} \label{negatives}
We devise object detection as language grounding of an utterance formed by concatenating a sequence of category labels, e.g. ``Chair. Dining table. Bed. Plant. Sofa.''. The task is again to i) detect the mentioned objects in the scene, i.e. return the bounding boxes of their instances, and ii) associate each localized box to a span, i.e. an object category in the utterance.

To form these detection prompts, one solution could be to concatenate all object classes into a long utterance. However, this can be impractical if the domain-vocabulary is ``open'', or, in practice, very large (485 classes in ScanNet, 1600 in Visual Genome and so on). Instead, assuming that we have object annotations, we sample out of the positive labels that are annotated for a scene and a number of negative ones, corresponding to class names that are not associated with any instances in the scene. Having negative classes in the detection prompts helps the precision of the model, as it learns not to fire for every noun phrase that appears in an utterance. More specifically, the contrastive losses described in the main paper push the negative class' text representation away from the query representation of existing objects.

MDETR also considers an object detection evaluation. However, there are two noticeable differences. First, they use only single-category utterances, e.g. ``Dog.''. This category can be either positive (appears in the annotations) or negative (does not appear in the annotations), according to a sampling ratio. Opposite to that, our detection prompts are longer, consisting of multiple object categories, both positive and negative. Second, MDETR employs these sentences after pre-training, to train and evaluate their model as an object detector. Instead, we mix detection prompts through the training, leading to considerable quantitative gains in both 3D and 2D.

Lastly, although the ratio $r$ of positive to negative classes that appear in a detection phrase is a hyperparameter, we report results only for $r=1$ and sample at most 10 positive classes. We leave tuning of this hyperparameter for future research.

\subsection{Detection results} \label{det_results}
A benefit of i) being able to ground all objects mentioned in the phrase and not only the target object, as well as ii) being trained with detection prompts, is that \model can operate as an object detector. We evaluate its performance on ScanNet benchmark which has 18 classes. Specifically, for each scene, we form a detection prompt that contains all 18 classes. The objective is to find all instances in the scene, as explained in Section \ref{negatives}.

We first train \model on ScanNet using the same prompt of 18 classes. This is analogous to a 3DETR \cite{Misra2021AnET} model with PointNet++ backbone or the DETR+KPS+iter ablation in Table 10 of \cite{Liu2021GroupFree3O}. Additionally, we evaluate BUTD-DETR trained on a language grounding benchmark. The results are shown in Table~\ref{table:scannet_results}. \model performs on par with the ablation of \cite{Liu2021GroupFree3O}, but worse than 3DETR. Note that our objectives, i.e. contrastive losses, are not optimized for classification across a fixed number of classes, but for query-span alignment. Instead, detectors use softmax layers over a known number of classes. For comparison, we train \model on ScanNet with a softmax loss over the 18 benchmark classes to observe an improvement of 1.7\%. However, softmax losses are not suitable for language grounding, where the labels are not a priori known or limited to a specific set. When \model is trained on the 3D referential datasets, the performance on ScanNet improves up to 3.7\%, without having access to more scenes. This suggests that co-training with grounding and detection prompts is beneficial for both tasks.
\begin{table}[t]
    \centering
    \caption{\textbf{Object detection performance on ScanNet.} We evaluate \model trained with detection prompts on different datasets. Training on referential data and detection prompts offers a consistent gain on detection mAP.}
    {
    \begin{tabular}{|c|c|}
        \hline
        Method & mAP@0.25 \\
        \hline
        DETR+KPS+iter \cite{Liu2021GroupFree3O} & 59.9 \\
        3DETR with PointNet++ \cite{Misra2021AnET} & 61.7 \\
        \hline
        \model trained on ScanNet & 59.3\\
        \model trained on ScanNet with softmax & 61.0\\
        \model trained on SR3D & 61.1 \\
        \model trained on NR3D & 61.3 \\
        \model trained on ScanRefer & \textbf{63.0} \\
        \hline
    \end{tabular}}
    \label{table:scannet_results}
\end{table}

\subsection{More qualitative results}
\label{qual}
\begin{figure*}[ht!]
	\centering
	\includegraphics[width=\textwidth]{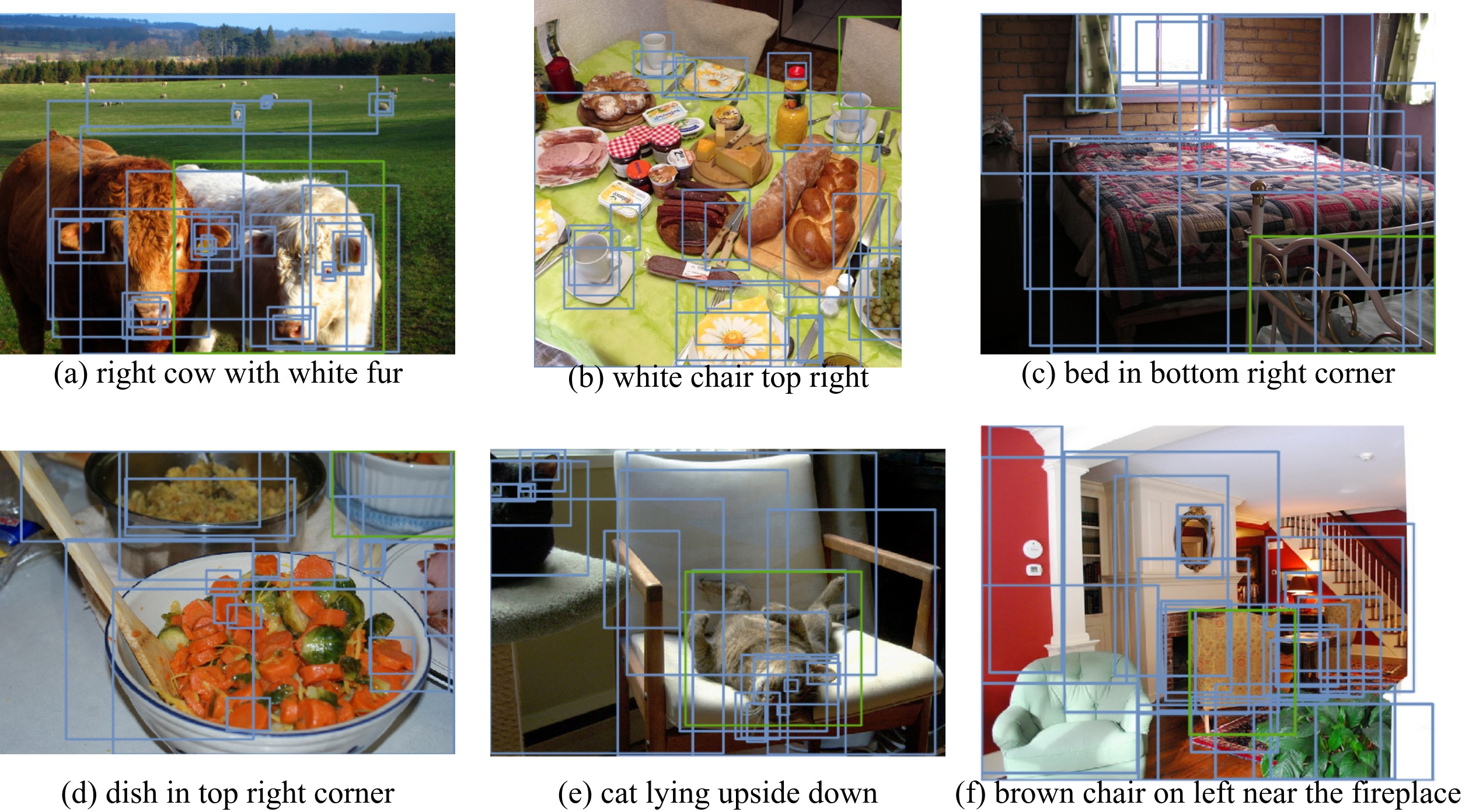}
	\caption{\textbf{Qualitative results of \model on RefCOCO.} The detector's proposals are shown in blue, our model's prediction in green. \model can predict boxes that the detector misses, e.g. in (b), the chair is missed by the detector so none of the previous detection-bottlenecked approaches could ground this phrase. In (a) and (c) the detector succeeds with low IoU but \model is able to predict a tight box around the referent object.}
	\label{fig:qualitative2d}
\end{figure*}

\begin{figure*}[t]
\begin{center}
    \includegraphics[width=\textwidth]{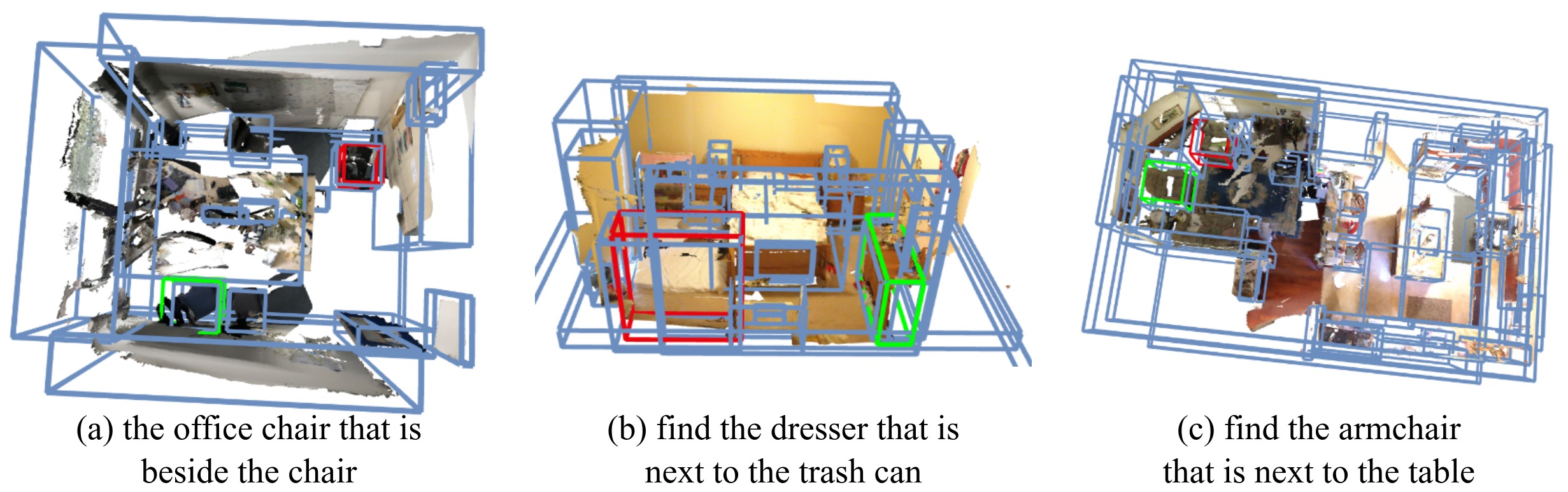}
\end{center}
\caption{Failure cases of \model on SR3D. Our predictions with red, ground-truth with green. Even if the box is there, still our model can fail, proving that ranking the correct boxes over other proposals remains a hard problem.}
\label{fig:suppl3d}
\end{figure*}

\begin{figure*}[t]
\begin{center}
    \includegraphics[width=\textwidth]{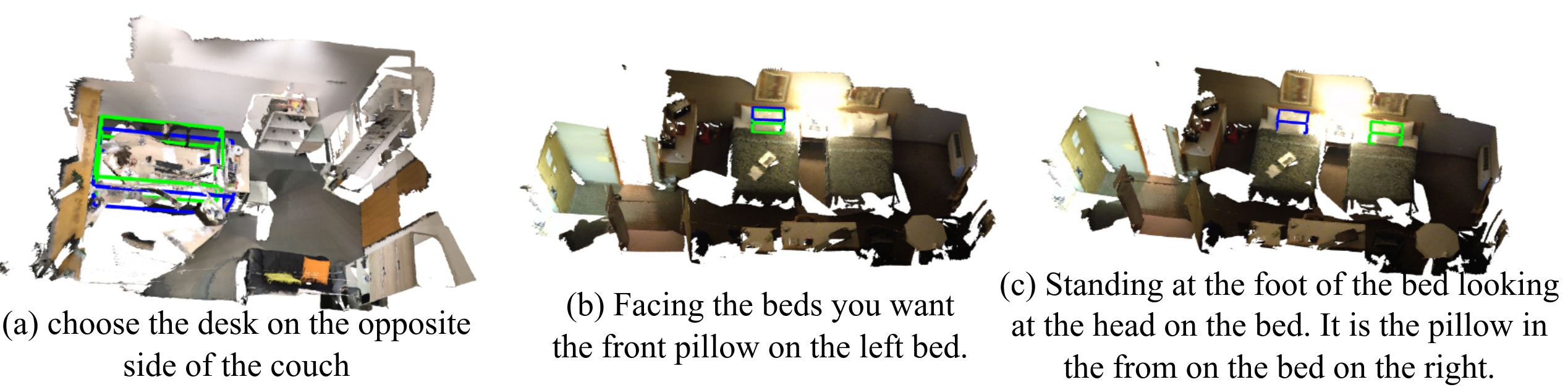}
\end{center}
\caption{Qualitative results of \model on NR3D. Our predictions are shown blue, ground-truth in green. The language of NR3D is more complex and the utterances are longer. Case (c) is a failure case.}
\label{fig:supplnr3d}
\end{figure*}

\begin{figure*}[t]
\begin{center}
    \includegraphics[width=\textwidth]{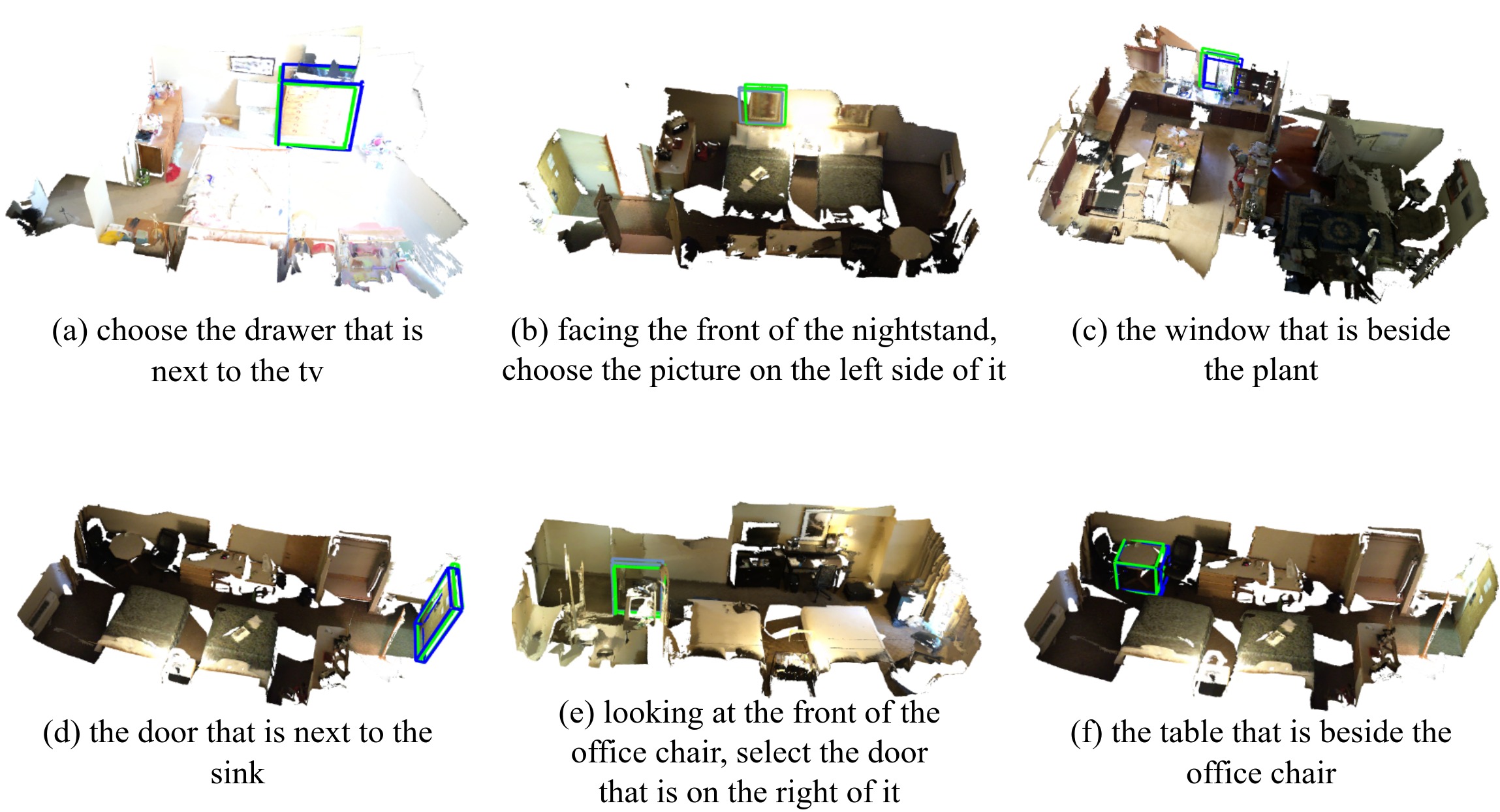}
\end{center}
\caption{More qualitative results of \model on SR3D. Our predictions are shown in blue, ground-truth in green.}
\label{fig:suppl3dm1}
\end{figure*}

\begin{figure*}[t]
\begin{center}
    \includegraphics[width=\textwidth]{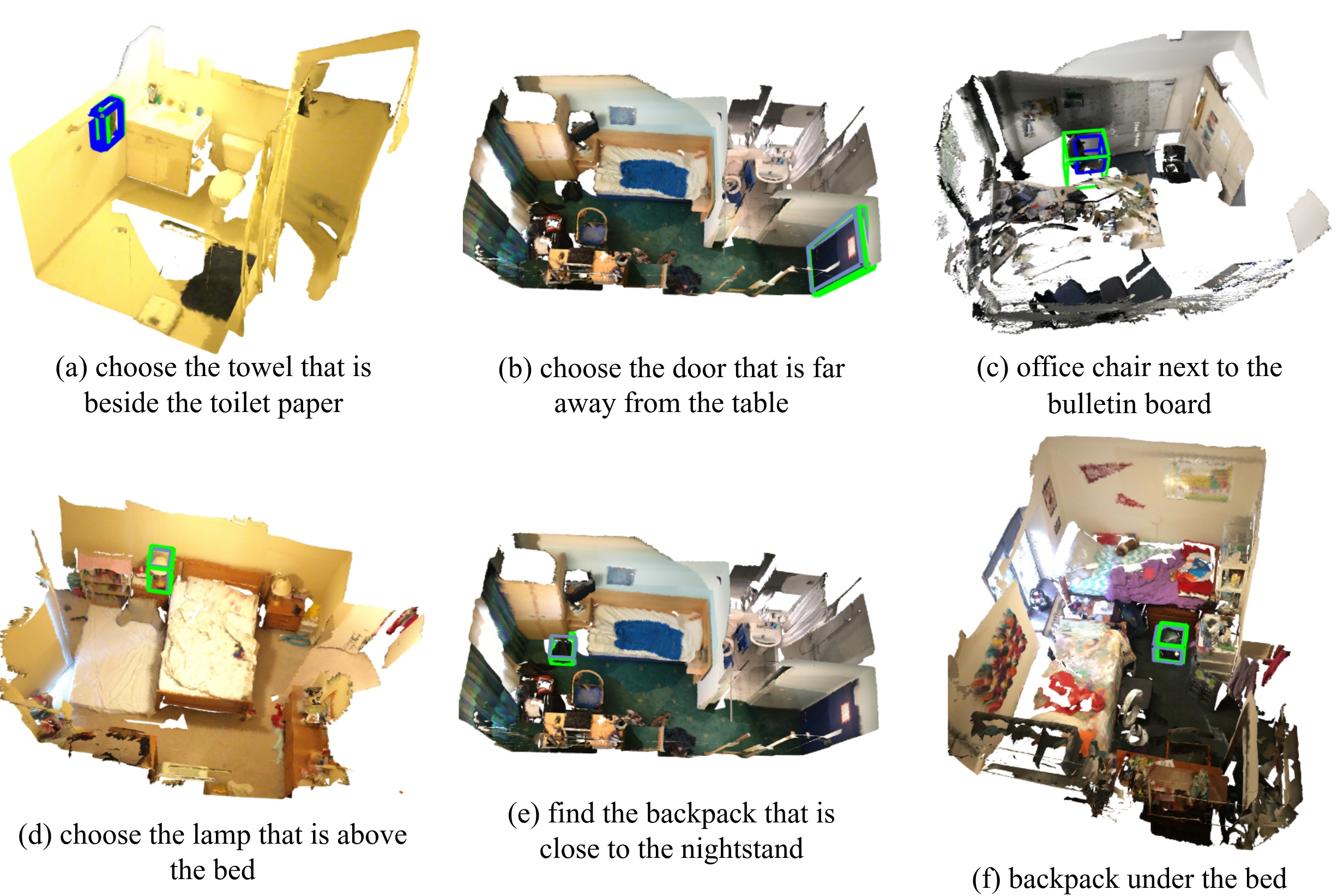}
\end{center}
\caption{More qualitative results of \model on SR3D. Our predictions are shown in blue, ground-truth in green.}
\label{fig:suppl3dm2}
\end{figure*}

We show qualitative results of the 2D version of \model on RefCOCO in Figure~\ref{fig:qualitative2d}. We also show failure cases on SR3D in Figure~\ref{fig:suppl3d}. More qualitative results on SR3D and NR3D are shown in Figures~\ref{fig:supplnr3d}, \ref{fig:suppl3dm1}, \ref{fig:suppl3dm2}.

\end{document}